\pdfoutput=1
\documentclass[letterpaper]{article} \usepackage{arxiv}  \usepackage[hyphens]{url}  \usepackage{graphicx}    \usepackage{natbib}  \usepackage{caption}   \usepackage{algorithm}
\usepackage{algorithmic}

\usepackage{amsmath}
\usepackage{amssymb}
\usepackage{makecell}
\usepackage{multirow}
\usepackage{placeins}
\usepackage{needspace}
\usepackage[hidelinks]{hyperref}
\usepackage{cleveref}
\newcommand{\NA}{--}

\usepackage{newfloat}
\usepackage{listings}
\DeclareCaptionStyle{ruled}{labelfont=normalfont,labelsep=colon,strut=off} 
\floatstyle{ruled}
\newfloat{listing}{tb}{lst}{}
\floatname{listing}{Listing}

\lstdefinestyle{prompt}{
    basicstyle=\fontsize{7pt}{7.5pt}\selectfont\ttfamily,
    numbers=none,
    frame=single,
    framesep=3pt,
    xleftmargin=0pt,
    columns=fullflexible,
    keepspaces=true,
    breaklines=true,
    breakindent=0pt,
    breakautoindent=false,
    showstringspaces=false,
    tabsize=2,
    aboveskip=1pt,
    belowskip=2pt
}
\floatstyle{ruled}
\newfloat{listing}{tb}{lst}{}
\floatname{listing}{Listing}
\newcommand{\promptcaption}[2]{            \Needspace{3\baselineskip}    \par\addvspace{2pt}    \refstepcounter{listing}\label{#2}    {\small\noindent\textbf{Listing \thelisting:} #1\par}    \nopagebreak[4]
}

\newenvironment{promptsingle}[2]{    \promptcaption{#1}{#2}}{    \par
}


\usepackage{booktabs}

\title{Multi-Agent Forensic Reasoning for Generalizable Deepfake Video Detection}
\author{
    \small
    Xuechao Zou\textsuperscript{\rm 1},
    Shun Zhang\textsuperscript{\rm 1},
    Kai Li\textsuperscript{\rm 2},
    Yi Zhou\textsuperscript{\rm 1},
    Xinyu Sun\textsuperscript{\rm 1},
    Yuhui Chen\textsuperscript{\rm 3},
    Zhe Wu\textsuperscript{\rm 2},
    Congyan Lang\textsuperscript{\rm 1},
    Junliang Xing\textsuperscript{\rm 2}
}
\affiliations{
    \textsuperscript{\rm 1}Beijing Jiaotong University \quad
    \textsuperscript{\rm 2}Tsinghua University \quad
    \textsuperscript{\rm 3}Ant Group
}
\copyrighttext{Project page: \url{https://xavierjiezou.github.io/ARGUS/}}
\begin{document}

\maketitle

\begin{abstract}

The malicious use of generative artificial intelligence to create highly realistic deepfake videos raises serious ethical concerns and poses substantial challenges to AI safety. However, existing deepfake video benchmarks provide limited coverage of recent synthesis methods and generally lack reliable fine-grained textual annotations. Meanwhile, conventional detectors and multimodal large language models (MLLMs), whether operating as a single model or relying on a single analytical perspective, often fail to capture subtle forgery artifacts, limiting their generalization to emerging AI-generated methods. To address these limitations, we introduce FaceVid-Forensics-100K, a large-scale deepfake video dataset comprising 100,000 videos and spanning 33 synthesis methods across face swapping, face reenactment, and entire-face synthesis, including recent generators such as Seedance 2.0. The dataset provides fine-grained textual annotations of visual observations and verdict-consistent forensic explanations, automatically synthesized through a multi-model aggregation and conflict-resolution pipeline powered by advanced MLLMs. Building on this benchmark, we propose a multi-agent forensic reasoning framework that employs four specialized domain-expert agents to independently analyze forgery cues from four perspectives: texture, lighting, motion, and physics. A judge agent then reconciles their reports to produce a final prediction together with an explanation. Extensive evaluations on out-of-domain test sets show that, despite being composed entirely of small open-source MLLMs, our framework outperforms all methods including closed-source GPT and Gemini models and ranks first across all reported metrics on this benchmark.
\end{abstract}

\begin{figure}[!t]
    \centering
    \includegraphics[width=\columnwidth]{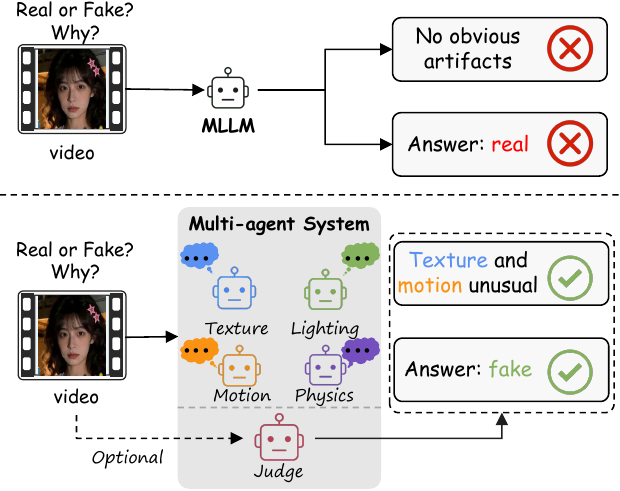}
    \caption{Framework comparison. A single MLLM often overlooks subtle forensic artifacts, which can lead to incorrect predictions. In contrast, our multi-agent framework employs specialized agents to examine the input video from four distinct forensic perspectives: texture, lighting, motion, and physics. A judge agent then aggregates their findings to produce a more reliable explanation and final prediction of whether the video under analysis is ultimately real or fake.}
    \label{fig:motivation}
\end{figure}

\section{Introduction}

The rapid advancement of generative artificial intelligence has made it increasingly easy to synthesize facial videos with high visual fidelity and temporal coherence. This progress creates unprecedented opportunities for creative industries, but also raises serious ethical and security concerns~\cite{li2024safeear}. Therefore, reliable deepfake video detection has become a critical task in digital media forensics. Yet the problem is no longer limited to identifying conspicuous blending boundaries or low-level generation artifacts. A new generation of video generators~\cite{seedance2.0,sora2,kling,wan2.1-t2v-1.3b} can synthesize entire faces, maintain relatively stable identity and appearance across consecutive frames, and produce seemingly plausible motion under diverse scenes. Consequently, detectors trained on earlier forgery techniques often suffer substantial performance degradation when confronted with unseen generation methods~\cite{cheng2026infodense,shen2025generative}.

Conventional detectors~\cite{altfreezing,tall++,tfcu,dfd-cfg,dfgaze,effort} are typically small vision models that learn discriminative visual representations primarily from binary real-or-fake labels. Although these methods can achieve strong in-domain performance, they often rely on shortcut features specific to particular datasets or generators and therefore struggle to adapt to new forgery methods, identities, compression conditions, and data sources~\cite{Huang_2026_CVPR}. Existing benchmarks~\cite{ff++,celeb-df,dfdc,deeperforensics,df40,celeb-df++} further amplify this problem: widely used datasets cover only a limited portion of the rapidly expanding synthesis landscape, and even some recent datasets that increase the number of videos typically provide only binary labels without specifying the visual evidence supporting each decision. Such coarse-grained supervision makes it difficult both to learn subtle, diverse, and previously unseen forgery cues and to determine whether a detector has acquired transferable forensic knowledge or merely fitted dataset biases. Moreover, authenticity decisions without supporting evidence limit the trustworthiness and auditability of detection systems in high-risk real-world settings~\cite{li2026audiotrust}.

Multimodal large language models (MLLMs)~\cite{VidGuard-R1,skyra,videoveritas,EDVD-LLaMA} offer a promising direction for addressing these limitations. With strong visual understanding and language-generation capabilities, MLLMs can analyze video content and describe suspicious facial details, inconsistencies across frames, and violations of common physical patterns. Nevertheless, directly asking a single MLLM whether a video is real or fake remains unreliable~\cite{longvideoagent,ma2026mmar}. A general-purpose model may overemphasize the most salient appearance cue, overlook weak evidence spread across different frames, or generate a plausible explanation that is inconsistent with its final verdict. More fundamentally, treating deepfake detection as a single holistic judgment ignores the distinct causes of different forgery traces: texture over-smoothing, inconsistent illumination, unstable motion, and violations of physical plausibility each require different forms of forensic knowledge and reasoning.

To address these limitations, we introduce the large-scale deepfake video benchmark FaceVid-Forensics-100K. The dataset contains 100,000 videos spanning 33 synthesis methods, including recent generators such as Seedance~2.0. It comprises 21,075 real and 78,925 fake videos across the major forgery categories of face swapping, face reenactment, and entire-face synthesis. Unlike existing datasets that provide only binary labels, FaceVid-Forensics-100K provides fine-grained textual supervision for every video along four dimensions: texture, lighting, motion, and physics. Specifically, multiple advanced open- and closed-source MLLMs first generate observations and judgments independently. An aggregation model then consolidates the evidence, resolves conflicts among the models, and produces a forensic explanation consistent with the final verdict. FaceVid-Forensics-100K therefore records not only authenticity labels but also the interpretable evidence supporting each decision, establishing a foundation for evidence-driven deepfake detection.

Building on this benchmark, we further propose a multi-agent forensic reasoning framework that decomposes deepfake detection into independent yet coordinated specialized analyses. Four domain-expert agents examine texture statistics, illumination consistency, temporal motion patterns, and physical plausibility, respectively. A judge agent then aggregates the expert reports, weighs mutually supporting or conflicting evidence, and outputs both an authenticity prediction and a concise forensic explanation. This decomposition encourages each agent to search systematically for a specific class of forgery traces, while the judge agent retains a global view of the video. When one type of forgery trace is weak or absent, the final decision can still rely on evidence corroborated across multiple perspectives, reducing the risk that a single salient cue dominates an erroneous prediction.

Extensive out-of-domain evaluations validate the effectiveness of the proposed approach. On the reported benchmark, the full system achieves 69.87\% accuracy, 81.82\% recall, and 53.28\% F1, outperforming small vision models, general-purpose open- and closed-source MLLMs, and forensics-tuned MLLMs. Compared with the strongest single-model baseline, our approach improves F1 from 47.45\% to 53.28\%, an absolute gain of 5.83 percentage points. Allowing the judge agent to access the video directly further increases F1 from 51.01\% to 53.28\%, indicating that the original visual information can effectively supplement the expert reports. These results show that explicit multi-perspective collaborative reasoning provides a more reliable basis for generalizable deepfake video detection than holistic judgment by a single MLLM.

\begin{table*}[t!]
\centering
\setlength{\tabcolsep}{0pt}
\renewcommand{\arraystretch}{0.95}

\begin{tabular*}{\textwidth}{@{\extracolsep{\fill}}l c c c c c c c@{}}
\toprule

 & \multicolumn{2}{c}{\textbf{Forgery Coverage}}
& \multicolumn{3}{c}{\textbf{Video Scale}}
& \multicolumn{2}{c}{\textbf{Textual Labels}} \\

\cmidrule(lr){2-3}
\cmidrule(lr){4-6}
\cmidrule(lr){7-8}

\multirow{-2}{*}[0.6ex]{\textbf{Dataset}}
&
\makecell[c]{\textbf{\#Synth.}\\ \textbf{Methods}}
&
\makecell[c]{\textbf{Latest}\\ \textbf{Fake}}
&
\textbf{Real}
&
\textbf{Fake}
&
\textbf{Total}
&
\textbf{Obs.}
&
\textbf{Exp.} \\

\midrule

DeepfakeDetection~\cite{deepfakedetection}
& 5
& \NA
& 363
& 3,068
& 3,431
& $\times$
& $\times$ \\

Celeb-DF v2~\cite{celeb-df}
& 1
& VAE (2014)
& 590
& 5,639
& 6,229
& $\times$
& $\times$ \\

DeeperForensics-1.0~\cite{deeperforensics}
& 1
& DF-VAE (2020)
& \textbf{50,000}
& 10,000
& 60,000
& $\times$
& $\times$ \\

DF40~\cite{df40}\textsuperscript{\textdagger}
& 23
& HeyGen (2024)
& 716
& 28,837
& 29,553
& $\times$
& $\times$ \\

Celeb-DF++~\cite{celeb-df++}
& 22
& FLOAT (2025)
& 590
& 53,196
& 53,786
& $\times$
& $\times$ \\

\midrule

\textbf{FaceVid-Forensics-100K}
& \textbf{33}
& \textbf{\mbox{Seedance 2.0 (2026)}}
& 21,075
& \textbf{78,925}
& \textbf{100,000}
& $\boldsymbol{\checkmark}$
& $\boldsymbol{\checkmark}$ \\

\bottomrule
\end{tabular*}

\caption{
Comparison of representative deepfake video datasets. \textsuperscript{\textdagger} Only the video-based subsets of DF40 are counted. Obs. and Exp. stand for Observation and Explanation, respectively.
}
\label{tab:dataset-comparison}
\end{table*}

\section{Related Work}

\subsection{Deepfake Video Detection}

Early deepfake benchmarks primarily focused on a small number of face-swapping~\cite{FaceShifter} and reenactment methods~\cite{thies2016face2face,NeuralTextures}. FaceForensics++~\cite{ff++} established a standardized benchmark across several manipulation pipelines, Celeb-DF~\cite{celeb-df} introduced higher-quality face swaps, and DFDC~\cite{dfdc} substantially increased the number of subjects and videos. DeeperForensics-1.0~\cite{deeperforensics} further incorporated real-world perturbations to evaluate robustness. As synthesis techniques~\cite{face-mogle,cao2026multivariate} diversified, DF40~\cite{df40} and Celeb-DF++~\cite{celeb-df++} broadened coverage across multiple face-forgery paradigms. These face-centric datasets have driven progress in scale, realism, and manipulation diversity, but their supervision remains predominantly binary. They therefore provide limited guidance about which visual evidence supports an authenticity decision. In contrast, FaceVid-Forensics-100K contains 100,000 face-centric videos spanning 33 synthesis methods, including recent systems such as Seedance~2.0, and augments binary authenticity labels with dimension-specific forensic observations and verdict-consistent explanations.

Most conventional detectors learn discriminative representations from binary labels. Some methods target manipulation traces such as blending boundaries~\cite{Face-X-Ray}, gaze behavior~\cite{dfgaze}, or lip motion~\cite{Lips-Donot-Lie}; others model temporal coherence and spatiotemporal inconsistency~\cite{9710282,altfreezing,tall++,tfcu}. Recent approaches improve transfer by adapting foundation-model features or suppressing generator-specific directions~\cite{dfd-cfg,effort,Cheng_2026_CVPR}. Despite these advances, such small vision models rely on binary supervision and can overfit specific artifacts, limiting generalization.

MLLMs make it possible to formulate deepfake detection as evidence-grounded visual reasoning rather than opaque binary classification. EDVD-LLaMA~\cite{EDVD-LLaMA} adapts an MLLM to explain manipulated facial videos, VidGuard-R1~\cite{VidGuard-R1} jointly improves detection and explanation through reinforcement learning, Skyra~\cite{skyra} grounds reasoning in annotated visual artifacts, and VideoVeritas~\cite{videoveritas} combines question--answer supervision with preference and perception-oriented reinforcement learning. However, single-model inference may overlook weak cues or let one artifact bias the verdict. Our framework instead enables explicit multi-perspective collaborative reasoning: four specialized agents produce dimension-specific evidence, which a judge reconciles into the final explanation and verdict.
\subsection{Multi-Agent Systems}

Multi-agent systems coordinate specialized decision makers through communication and information exchange. Recent LLM-based systems use debate, critique, and iterative collaboration to improve factuality and reasoning~\cite{du2024improving,wu2024autogen,gao2025singleagentmultiagentsystemsboth,he2025enhancingllmreasoningmultipath}, while learning-based approaches optimize the interaction policies of collaborating agents~\cite{MARPO,drmas,PettingLLMs,qiao2026offline}. Multi-agent designs have also been extended to multimodal tasks: LongVideoAgent~\cite{longvideoagent} coordinates grounding and visual agents for long-video understanding, and UniShield~\cite{Unishield} routes specialized forensic tools for image manipulation detection and localization. Inspired by role-specialized multi-agent collaboration, we decompose deepfake video detection into four independently forensic perspectives, with each agent generating observation evidence. A judge then reconciles the evidence, establishing an explicit multi-perspective collaboration mechanism.

\section{FaceVid-Forensics-100K}

\subsection{Dataset Overview}

\begin{figure*}[t!]
    \centering
    \includegraphics[width=0.90\textwidth]{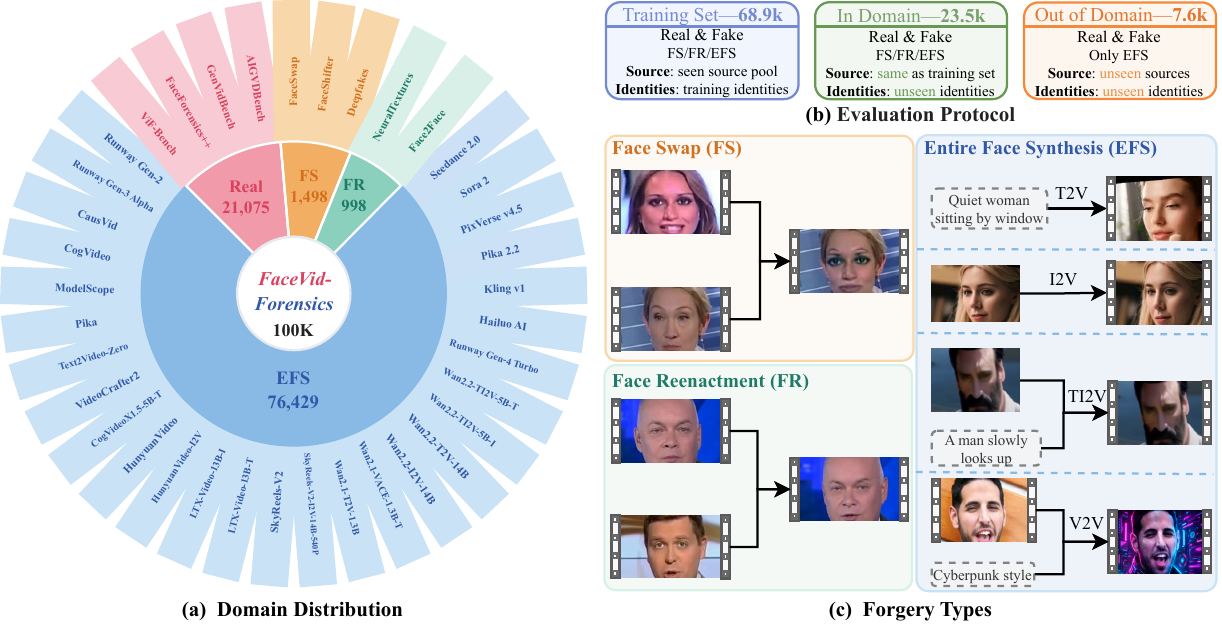}
    \caption{Overview of FaceVid-Forensics-100K.}
    \label{fig:wide}
\end{figure*}

FaceVid-Forensics-100K is a large-scale deepfake video dataset comprising 100,000 face-centric videos. It covers 33 synthesis methods across major forgery categories, including face swapping, face reenactment, and entire-face synthesis. Beyond binary authenticity labels, the dataset provides fine-grained textual annotations of visual observations across four forensic dimensions---texture, lighting, motion, and physics---as well as verdict-consistent explanations. As shown in~\Cref{fig:wide}, it serves as a foundation for training and evaluating evidence-driven deepfake detection systems.

\paragraph{Collection and Processing.}
Our dataset originates from two types of sources: existing general video forgery datasets and videos directly collected from the internet or synthesized via recent generative models. Specifically, we collect videos from AIGVDBench~\cite{AIGVDBench}, GenVidBench~\cite{genvidbench}, ViF-Bench~\cite{skyra}, and FaceForensics++ (FF++)~\cite{ff++}, yielding approximately $442{,}000$, $6{,}780{,}000$, $3{,}000$, and $5{,}000$ videos, respectively. In addition, we crawl $2{,}901$ videos generated by Seedance 2.0~\cite{seedance2.0} and other recent models from the internet, retaining $577$ videos after manual screening. Following data collection, we apply a face detection pipeline to filter the videos frame-by-frame, retaining only frames containing faces and discarding videos without valid facial regions. After this preprocessing step, we obtain $1{,}158{,}585$ valid face-centric videos, comprising $30{,}240$ real videos and $1{,}128{,}345$ fake videos, with a total duration of $436.56$ hours.

To ensure a balanced distribution, we first deduplicate the real videos based on their YouTube IDs, which reduces the real samples from $30{,}240$ to $21{,}075$. For the fake videos, four synthesis models (ModelScope~\cite{modelscope}, Pika~\cite{pika}, Text2Video-Zero~\cite{text2videozero}, and VideoCrafter2~\cite{videocrafter2}) dominate the collection. We evaluate these fake videos using the AltFreezing~\cite{altfreezing} detector and prioritize retaining samples with lower scores (i.e., those harder to distinguish). Through this strategy, the number of videos from these four models is reduced from $235{,}007$, $75{,}471$, $435{,}210$, and $376{,}471$ to $11{,}958$, $15{,}163$, $10{,}432$, and $32{,}690$, respectively. This decreases their total count from $1{,}122{,}159$ to $70{,}243$. Combined with the other retained fake videos, the total number of fake videos is reduced from $1{,}128{,}345$ to $78{,}925$. The final curated dataset contains $100{,}000$ videos, consisting of $21{,}075$ real videos and $78{,}925$ fake videos. These forgeries cover face swapping (FS), face reenactment (FR), and entire-face synthesis (EFS), which encompasses text-to-video (T2V), image-to-video (I2V), text-and-image-to-video (TI2V), and video-to-video (V2V) generation paradigms.

\paragraph{Training and Evaluation.}
We divide the dataset into training, in-domain test, and out-of-distribution (OOD) test sets. The training and in-domain test sets share the same families of synthesis methods. To rigorously evaluate cross-generator generalization, we assign videos from $20$ completely unseen (held-out) EFS generators exclusively to the OOD test set, along with a portion of the real videos. As detailed in Appendix~\ref{sec:dataset}, these held-out generators include recent systems such as CogVideoX1.5, HunyuanVideo, LTX-Video, SkyReels, Wan, Hailuo, Kling, Pika 2.2, PixVerse, Sora 2, and Seedance 2.0. In total, the training, in-domain test, and OOD test splits contain approximately 68.9K, 23.5K, and 7.6K videos, respectively.

\begin{figure*}[t!]
    \centering
    \includegraphics[width=0.90\textwidth]{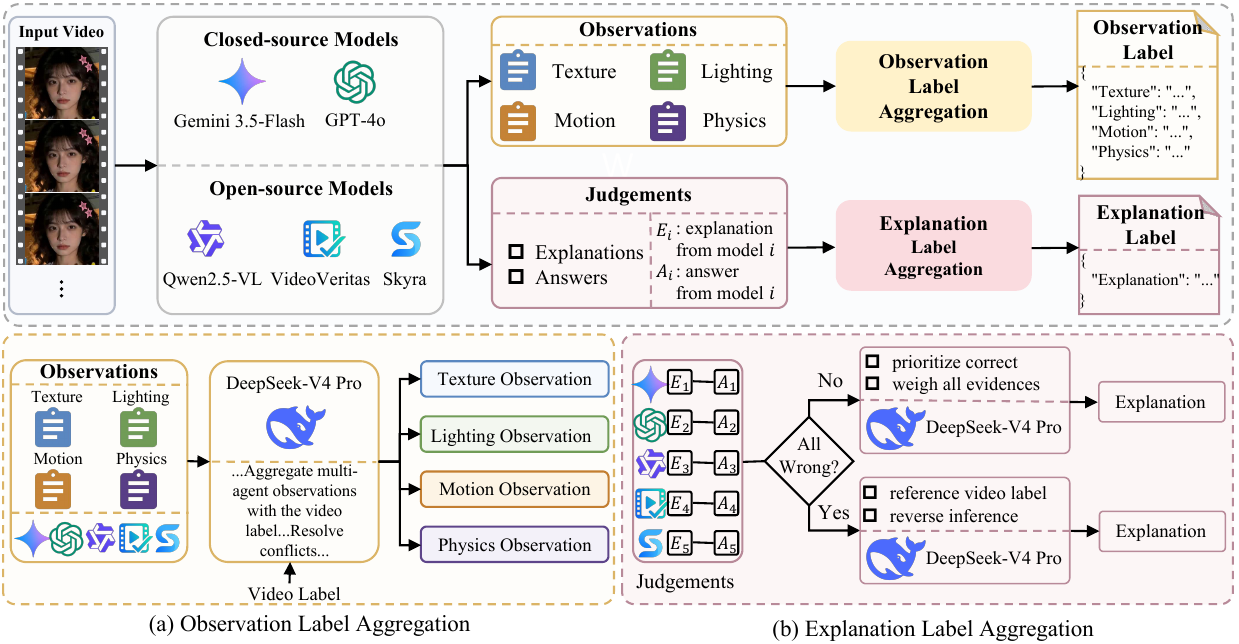}
    \caption{Pipeline of observation and explanation label generation. (a) DeepSeek-V4 Pro aggregates independent observations separately within each forensic dimension. (b) It combines answer--explanation pairs into a verdict-consistent explanation label.}
    \label{label_generation}
\end{figure*}

\subsection{Label Generation}

To construct fine-grained textual supervision without relying on manual annotation, we employ an ensemble of five diverse MLLMs spanning three distinct paradigm types as annotators, as shown in~\Cref{label_generation}: (1)~powerful closed-source MLLMs (GPT-4o~\cite{gpt4o} and Gemini 3.5-Flash~\cite{gemini35flash}); (2)~general-purpose open-source MLLMs (Qwen2.5-VL~\cite{qwen25vl}); and (3)~forensics-tuned domain-specific MLLMs (Skyra~\cite{skyra} and VideoVeritas~\cite{videoveritas}). To synthesize their outputs into unified, high-quality textual supervision, we select DeepSeek-V4 Pro~\cite{deepseek-v4} as the central aggregator based on two key design considerations:
first, label aggregation operates strictly over textual observation/explanation reports and ground-truth metadata, making a pure text LLM with strong logical reasoning capabilities optimal without requiring multimodal visual inputs;
second, decoupling the aggregator from the annotator pool ensures architectural independence, preventing the aggregator from inheriting potential inductive biases or error patterns present in the visual annotator MLLMs.

\paragraph{Observation Label.}
As illustrated in~\Cref{label_generation}(a), the five annotators independently analyze each video to generate detailed observations across the four forensic dimensions (Texture, Lighting, Motion, and Physics). The aggregator model (DeepSeek-V4 Pro) then integrates these multi-annotator outputs by merging the observations separately within each forensic dimension. Guided by the ground-truth video label to resolve cross-model contradictions and filter noise, the aggregator outputs a structured observation label containing consensus descriptions for each dimension without revealing the final authenticity verdict.

\paragraph{Explanation Label.}
In addition to dimensional observations, each annotator produces an authenticity verdict (real or fake) alongside an initial explanation rationale. As shown in~\Cref{label_generation}(b), the aggregator model consolidates these individual answer--explanation pairs into a single, verdict-consistent explanation label. This aggregation is strictly conditioned on the ground-truth video label: when correct predictions exist, the aggregator prioritizes and synthesizes explanations from correct annotators; if all annotators make incorrect predictions, it performs reverse inference by re-evaluating the merged observations against the ground-truth label. This ensures the final textual rationale is aligned with the correct authenticity verdict while eliminating individual model biases, noise, and hallucinations.

\section{Multi-Agent Forensic Reasoning}

\begin{figure*}[t!]
    \centering
    \includegraphics[width=0.90\textwidth]{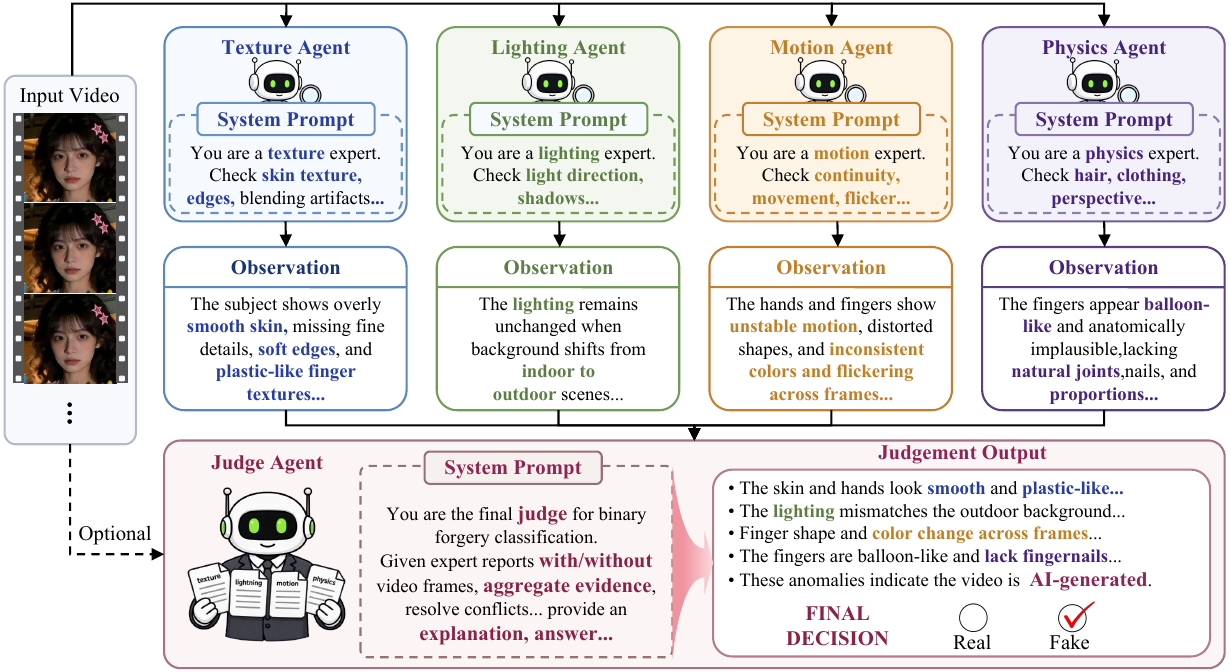}
    \caption{Overview of the proposed multi-agent forensic reasoning framework.}
    \label{method_pipeline}
\end{figure*}

\subsection{Multi-Agent System}

As shown in~\Cref{method_pipeline}, our framework decomposes deepfake video detection into independent forensic analyses coordinated by a hierarchical multi-agent system. For a given input video $v$, we uniformly sample a sequence of frames $X(v)$ and provide them to the system, which consists of four specialized observation agents and a central judge agent.

\paragraph{Observation Agents.} 
The four observation agents focus on distinct aspects of forgery-related artifacts: texture, lighting, motion, and physics. Each observation agent $\mathcal{A}_d$ independently receives the same sampled frames $X(v)$ along with a dimension-specific prompt, and produces a textual observation $\widehat{O}^{d}$ without predicting the final authenticity. We instantiate all agents using MLLMs. Their system prompts instruct each agent to search systematically for its assigned cue type: the texture agent evaluates skin details and blending boundaries, the lighting agent assesses illumination and shadow consistencies, the motion agent tracks temporal instability, and the physics agent considers anatomical and geometric plausibility. The complete prompts for these agents are detailed in Appendix~\ref{sec:prompt-templates}.

\paragraph{Judge Agent.} 
The judge agent acts as the central reasoning hub. It receives the textual observations generated by the four observation agents and, optionally, the sampled video frames $X(v)$. Its prompt instructs it to weigh mutually supporting or conflicting evidence from the experts and output a binary real-or-fake answer alongside a concise, verdict-consistent explanation. This explicit multi-perspective collaboration ensures that decisions rely on corroborated evidence rather than being overly influenced by a single salient artifact. The exact prompt template is provided in Appendix~\ref{sec:prompt-templates}.

\subsection{Training Pipeline}

We train the proposed multi-agent framework in two sequential stages: supervised fine-tuning (SFT) for all agents, followed by group relative policy optimization (GRPO)~\cite{grpo} to refine the judge agent's decisions.

\paragraph{Supervised Fine-Tuning.}
During SFT, the four observation agents are trained independently using the aggregated, dimension-specific observation labels $O_\star^d=(o_1^d,\ldots,o_{L_d}^d)$ collected during the dataset construction process. For each dimension $d \in \mathcal{D}$, the corresponding agent $\mathcal{A}_d$ is optimized via autoregressive language modeling:
\begin{equation}
\mathcal{L}_{\mathrm{obs}}^d(\theta_d)
=-\frac{1}{N_d}
\sum_{(v,O_\star^d)\in\mathcal{T}}
\sum_{t=1}^{L_d} 
\log p_{\theta_d}\!\left(
o_t^d \mid X(v),I_d,o_{<t}^d
\right),
\end{equation}
where $\mathcal{T}$ is the training set, $\theta_d$ denotes the agent's parameters, $I_d$ is the instruction, and $N_d$ is the number of response tokens. 
After training the observation agents, we apply them to the training set to generate their intermediate outputs $R(v) = \operatorname{Concat}_{d\in\mathcal{D}}(\widehat{O}^{d})$. The judge agent is then trained on these concatenated observations and the visual input $V_c(v)$ (where $c\in\{0,1\}$ indicates the presence of video frames), taking $Q_c(v)=(R(v), V_c(v))$ as input. The judge agent learns to produce the target explanation and verdict $Z_\star=(z_1,\ldots,z_{L_J})$ by minimizing:
\begin{equation}
\mathcal{L}_{\mathrm{J}}^c(\phi_c)
=-\frac{1}{N_{\mathrm{J}}}
\sum_{(v,Z_\star)\in\mathcal{T}}
\sum_{t=1}^{L_J}
\log p_{\phi_c}\!\left(
z_t\mid Q_c(v),z_{<t}
\right).
\end{equation}

\paragraph{Policy Optimization for Decision Refinement.}
Following SFT, we apply GRPO to further align the judge agent's final decision policy, keeping the four observation agents frozen. For each input $Q_c(v)$, the judge agent samples a group of $G$ candidate responses $\{\widetilde{Z}_g\}_{g=1}^{G}$. A binary accuracy reward $r_g \in \{0, 1\}$ is assigned based on whether the candidate correctly predicts the ground-truth label. We compute the relative advantage $A_g$ by normalizing the rewards within the group:
\begin{equation}
A_g=\frac{r_g-\overline{r}}{\sigma_r+\epsilon}.
\end{equation}
GRPO then updates the judge agent's parameters to maximize this advantage using a clipped surrogate objective, complemented by a KL-divergence penalty against the frozen SFT policy to maintain explanation quality. This stage refines the final classification accuracy without altering the specialized evidence extracted by the observation agents.

\section{Experiments}

\subsection{Experimental Setup}

\paragraph{Implementation Details.} To ensure a fair comparison with Skyra~\cite{skyra}, we adopt Qwen2.5-VL-7B~\cite{qwen25vl} as the base model for all agents. All training is implemented using ms-swift~\cite{ms-swift}. We train each component for one epoch using LoRA~\cite{lora} and the AdamW optimizer~\cite{adamw}, with learning rates of $1\times10^{-4}$ for supervised fine-tuning and $5\times10^{-6}$ for GRPO. For GRPO, we sample eight responses per prompt.

\paragraph{Evaluation Protocols.} We conduct the evaluation on the OOD split set, containing 5,716 real videos and 1,920 fake videos from 20 generators excluded from training. For small vision models, we retrain each method on FaceVid-Forensics-100K using its official code and data preprocessing pipeline. Off-the-shelf general MLLMs and forensics-tuned MLLMs are evaluated using their released checkpoints or official APIs. For models that support an explicit thinking mode, we disable it during inference. Following Skyra, we use accuracy (Acc), F1 score, and recall as the evaluation metrics.

\subsection{Comparison with State-of-the-Art Methods}

As shown in~\Cref{tab:ood_overall}, we compare our framework with small vision models, general-purpose open- and closed-source MLLMs~\cite{qwen3.6-35b-a3b,mimov25,gpt4o,gemini35flash}, and forensics-tuned MLLMs on the OOD test set. Our framework achieves the strongest overall performance when the judge agent receives both the observation reports and sampled video frames. The text-only judge agent also ranks second, showing that the observation agents provide effective forensic evidence, while direct access to the video further improves generalization. In-domain results, per-generator OOD results, qualitative comparisons, and evaluation of textual explanation quality are available at Appendix~\ref{sec:additional-experimental-results}.

\begin{table}[t!]
\centering
\begin{tabular*}{\linewidth}{@{\extracolsep{\fill}}lccc@{}}
\toprule
\textbf{Method} & \textbf{Acc} & \textbf{Recall} & \textbf{F1} \\
\midrule
\multicolumn{4}{@{}l}{\textbf{Small Vision Models}} \\
DFGaze \textit{(TIFS'24)} & 56.67 & 17.92 & 27.24 \\
DFD-FCG \textit{(CVPR'25)} & 61.25 & 29.22 & 39.16 \\
Effort \textit{(ICML'25)} & 63.92 & 34.79 & 44.76 \\
TALL++ \textit{(IJCV'24)} & 63.47 & 42.97 & 45.07 \\
TFCU \textit{(CVPR'25)} & 64.28 & 33.44 & 45.20 \\
\midrule
\multicolumn{4}{@{}l}{\textbf{Open-source MLLMs}} \\
Qwen2.5-VL-7B & 35.94 & 16.51 & 13.24 \\
InternVL3.5-8B & 37.87 & 17.60 & 14.53 \\
MiMo-V2.5 (310B-A15B) & 49.85 & 15.00 & 18.69 \\
Qwen3.6-35B-A3B & 53.17 & 50.05 & 35.72 \\
\midrule
\multicolumn{4}{@{}l}{\textbf{Closed-source MLLMs}} \\
GPT-4o (2024) & 57.63 & 40.36 & 37.55 \\
GPT-5-mini (2025) & 59.31 & 38.70 & 39.00 \\
Gemini-2.5-Pro (2025) & 63.78 & 75.29 & 47.45 \\
Gemini-3.5-Flash (2026) & 63.34 & 58.75 & 46.22 \\
\midrule
\multicolumn{4}{@{}l}{\textbf{Forensics-tuned MLLMs}} \\
Skyra \textit{(CVPR'26)} & 60.50 & 30.05 & 38.30 \\
VideoVeritas \textit{(ICML'26)} & 57.87 & \underline{78.96} & 43.22 \\
\midrule
\multicolumn{4}{@{}l}{\textbf{Multi-Agent System}} \\
Ours (w/o Video) & \underline{67.41} & 65.00
& \underline{51.01} \\
Ours (w/ Video) & \textbf{69.87} & \textbf{81.82}
& \textbf{53.28} \\
\bottomrule
\end{tabular*}
\caption{Comparison with state-of-the-art deepfake video detectors on the OOD
test set. The best and
second-best available results are highlighted in bold and underlined.}
\label{tab:ood_overall}
\end{table}

\subsection{Ablation Studies}

\paragraph{Contribution of Each Observation Agent.}
As shown in~\Cref{tab:single-observation-ablation}, combining all four observation agents gives the best F1 score under both judge agent configurations. The advantage is clearer without direct video input, indicating that the four perspectives provide distinct evidence. When frames are also available, the gap narrows because the judge agent can recover part of the missing visual information directly.

\begin{table}[t!]
\centering
\setlength{\tabcolsep}{1.8pt}
\begin{tabular*}{\linewidth}{@{\extracolsep{\fill}}l ccc ccc@{}}
\toprule
 & \multicolumn{3}{c}{\textbf{Ours (w/o Video)}} & \multicolumn{3}{c}{\textbf{Ours (w/ Video)}} \\
\cmidrule(lr){2-4} \cmidrule(lr){5-7}
\multirow{-2}{*}[0.6ex]{\textbf{Observation}} & \textbf{Acc} & \textbf{Recall} & \textbf{F1} & \textbf{Acc} & \textbf{Recall} & \textbf{F1} \\
\midrule
Texture  & 58.39 & \textbf{97.40} & 44.53 & \underline{66.90} & \textbf{87.86} & \underline{50.39} \\
Lighting & \underline{59.60} & \underline{94.22} & \underline{45.13} & 65.87 & 85.16 & 49.49 \\
Motion   & 54.70 & 75.31 & 40.54 & 63.15 & 83.39 & 47.20 \\
Physics  & 56.43 & 91.82 & 43.02 & 62.85 & 87.45 & 47.11 \\
\midrule
All four & \textbf{63.61} & 82.24 & \textbf{47.53} & \textbf{67.03} & \underline{87.76} & \textbf{50.49} \\
\bottomrule
\end{tabular*}
\caption{Comparison of individual observation agent outputs and their combination on the OOD test set. The best and second-best results within each judge agent configuration are highlighted in bold and underlined, respectively.}
\label{tab:single-observation-ablation}
\end{table}

\paragraph{Effect of Training with SFT and GRPO.}
To isolate the contribution of each trained component, Only-Observation trains the observation agents while keeping the Judge untrained, whereas Only-Judge trains the Judge using outputs from untrained observation agents. As shown in~\Cref{tab:sft_visual_ablation}, training either component improves the training-free system, jointly training the observation agents and judge is more effective than training either alone, and GRPO provides a further gain. These results confirm that specialized observation learning, evidence reconciliation, and policy optimization each contribute to the final generalizable performance. Relevant training dynamics are provided in Appendix~\ref{sec:training-dynamics}. Furthermore, we investigate the impact of different combinations of MLLMs for the observation and judge agents in Appendix~\ref{sec:observation-judge-models} and the impact of model parameter scale in Appendix~\ref{sec:model-scale-ablation}.

\begin{table}[t!]
\centering
\setlength{\tabcolsep}{1.2pt}
\begin{tabular*}{\linewidth}{@{\extracolsep{\fill}}l ccc ccc@{}}
\toprule
 & \multicolumn{3}{c}{\textbf{w/o Video}} & \multicolumn{3}{c}{\textbf{w/ Video}} \\
\cmidrule(lr){2-4} \cmidrule(lr){5-7}
\multirow{-2}{*}[0.6ex]{\textbf{Stage}} & \textbf{Acc} & \textbf{Recall} & \textbf{F1} & \textbf{Acc} & \textbf{Recall} & \textbf{F1} \\
\midrule
Training-Free     & 45.06 & 65.05 & 33.53 & 42.29 & 54.06 & 30.07 \\
+SFT (Only-Obs)   & 50.63 & 99.58 & 40.46 & 50.64 & 99.84 & 40.48 \\
+SFT (Only-Judge) & 51.42 & 99.95 & 40.87 & 64.29 & 95.78 & 48.39 \\
+SFT (Joint)      & 63.61 & \underline{82.24} & 47.53 & 67.03 & \textbf{87.76} & 50.49 \\
+SFT+GRPO         & \underline{67.41} & 65.00 & \underline{51.01} & \textbf{69.87} & 81.82 & \textbf{53.28} \\
\bottomrule
\end{tabular*}
\caption{Effect of training with SFT and GRPO on the OOD test set. The best and second-best results are highlighted in bold and underlined, respectively.}
\label{tab:sft_visual_ablation}
\end{table}

\paragraph{Reasoning Strategies across MLLMs.}
We compare our multi-agent design with direct prediction, chain-of-thought (CoT) prompting, and three multi-turn variants using frozen MLLMs, as shown in~\Cref{tab:training_free_backbones}. Specifically, all multi-turn baselines decompose the forensic reasoning into 5 sequential dialogue turns (4 observation turns, followed by a final decision turn), but differ in frame delivery across turns: \textit{Multi-turn} inputs video frames only in the first turn, \textit{Multi-turn-Obs} provides frames across all four observation turns (turns 1--4), and \textit{Multi-turn-All} feeds frames continuously across all 5 turns. Across both backbones, our framework performs best overall. This result shows that the improvement comes from independent evidence collection and Judge-based reconciliation, rather than from longer prompts, additional dialogue turns, or repeated access to the video frames. Detailed implementations, efficiency analyses, and qualitative comparisons of these reasoning strategies are provided in Appendix~\ref{sec:reasoning-strategies}.

\begin{table}[t!]
\centering
\setlength{\tabcolsep}{1.8pt}
\begin{tabular*}{\linewidth}{@{\extracolsep{\fill}}l ccc ccc@{}}
\toprule
 & \multicolumn{3}{c}{\textbf{Qwen2.5-VL-7B}} & \multicolumn{3}{c}{\textbf{InternVL3.5-8B}} \\
\cmidrule(lr){2-4} \cmidrule(lr){5-7}
\multirow{-2}{*}[0.6ex]{\textbf{Strategy}} & \textbf{Acc} & \textbf{Recall} & \textbf{F1} & \textbf{Acc} & \textbf{Recall} & \textbf{F1} \\
\midrule
Single & 35.94 & 16.51 & 13.24 & 37.87 & 17.60 & 14.53 \\
CoT & 30.93 & 15.31 & 11.29 & 41.90 & 2.81 & 3.53 \\
Multi-turn & 33.98 & 15.26 & 11.92 & 40.53 & 41.56 & 25.85 \\
Multi-turn-Obs & 33.46 & 11.61 & 9.50 & 38.12 & 40.26 & 24.34 \\
Multi-turn-All & 33.88 & 13.87 & 11.05 & 37.22 & 38.80 & 23.48 \\
Ours (w/o Video) & \textbf{45.06} & \textbf{65.05} & \textbf{33.53} & \textbf{45.40} & \underline{46.46} & \underline{29.81} \\
Ours (w/ Video) & \underline{42.29} & \underline{54.06} & \underline{30.07} & \underline{45.29} & \textbf{49.22} & \textbf{30.52} \\
\bottomrule
\end{tabular*}
\caption{Training-free comparison of reasoning strategies across MLLMs on the OOD test set. Best and second-best results within each backbone are highlighted in bold and underlined, respectively.}
\label{tab:training_free_backbones}
\end{table}

\FloatBarrier
\section{Conclusion}

We presented FaceVid-Forensics-100K, a large-scale deepfake video benchmark with broad coverage of recent synthesis methods and fine-grained forensic annotations. We further proposed a multi-agent forensic reasoning framework that performs collaborative analysis from four complementary forensic perspectives and produces both authenticity predictions and explanations. Extensive experiments demonstrate that our approach consistently outperforms existing vision-based detectors and MLLMs on out-of-domain benchmarks, highlighting the effectiveness of multi-perspective collaborative reasoning for generalizable deepfake video detection.

\clearpage
\appendix
\setcounter{secnumdepth}{2}

\section{Additional Details of FaceVid-Forensics-100K}\label{sec:dataset}

\Cref{tab:dataset} summarizes the source- and method-level composition of the three splits. The real subset contains 21,075 videos: 12,934 from AIGVDBench~\cite{AIGVDBench}, 7,058 from GenVidBench~\cite{genvidbench}, 999 from FF++~\cite{ff++}, and 84 from ViF-Bench~\cite{skyra}. After preprocessing and deduplication, 11,626 real videos are assigned to training, 3,733 to in-domain testing, and 5,716 to out-of-distribution (OOD) testing. For conventional face manipulations, videos are drawn from five manipulation methods in FF++ and grouped by manipulation type. The face-swapping (FS) subset includes Deepfakes, FaceSwap, and FaceShifter, with 1,288 videos assigned to training and 210 to in-domain testing. The face-reenactment (FR) subset includes Face2Face and NeuralTextures, with 858 videos assigned to training and 140 to in-domain testing.

The entire-face synthesis (EFS) subset is split at the generator level. Eight seen generators contribute 74,509 videos: Runway Gen-2 and Runway Gen-3 Alpha~\cite{runwaygen3}, CausVid~\cite{causvid}, CogVideo~\cite{cogvideo}, ModelScope~\cite{modelscope}, Pika~\cite{pika}, Text2Video-Zero~\cite{text2videozero}, and VideoCrafter2~\cite{videocrafter2}. Of these, 55,134 are for training and 19,375 for in-domain testing, with counts ranging from 456 (Runway Gen-2) to 32,690 (VideoCrafter2). The OOD split contains 1,920 fake videos from 20 additional generators drawn from the CogVideoX, HunyuanVideo, LTX-Video, SkyReels, Wan, Runway, Hailuo, Kling, Pika, PixVerse, Sora, and Seedance families~\cite{cogvideox1.5-5b-t,hunyuanvideo,ltx-video-13b-I,skyreels-v2,wan2.1-t2v-1.3b,runwaygen4turbo,hailuo,kling,pika2.2,pixverse-v4.5,sora2,seedance2.0}, ranging from 34 to 576 per generator. Spanning text-to-video, image-to-video, text-and-image-to-video, and video-to-video settings, these generators are excluded from training and in-domain testing. This generator-disjoint split evaluates generalization to unseen synthesis methods.
\begin{table*}[!t]
\centering

\begin{tabular*}{\textwidth}{@{\extracolsep{\fill}}l l r r r r@{}}
\toprule
\textbf{Category}
& \textbf{Source}
& \textbf{Train}
& \makecell{\textbf{In-Domain}\\\textbf{Test}}
& \makecell{\textbf{Out-of-}\\\textbf{Distribution}}
& \textbf{Total} \\
\midrule

\multicolumn{6}{@{}c@{}}
{\textit{\textbf{Real Videos~--~21{,}075}}} \\
\midrule

\multirow{4}{*}{Real}
& AIGVDBench  & 9{,}706 & 3{,}228 & 0       & 12{,}934 \\
& GenVidBench & 1{,}061 & 365     & 5{,}632 & 7{,}058 \\
& FF++        & 859     & 140     & 0       & 999 \\
& ViF-Bench   & 0       & 0       & 84      & 84 \\

\midrule
\multicolumn{6}{@{}c@{}}
{\textit{\textbf{Fake Videos~--~78{,}925}}} \\
\midrule

\multirow{3}{*}{FS}
& Deepfakes   & 429 & 70 & 0 & 499 \\
& FaceSwap    & 430 & 70 & 0 & 500 \\
& FaceShifter & 429 & 70 & 0 & 499 \\

\cmidrule(lr){1-6}

\multirow{2}{*}{FR}
& Face2Face       & 429 & 70 & 0 & 499 \\
& NeuralTextures & 429 & 70 & 0 & 499 \\

\cmidrule(lr){1-6}

\multirow{28}{*}{EFS}
& Runway Gen-2             & 320     & 136     & 0   & 456 \\
& Runway Gen-3 Alpha       & 620     & 271     & 0   & 891 \\
& CausVid                  & 375     & 162     & 0   & 537 \\
& CogVideo                 & 1{,}787 & 595     & 0   & 2{,}382 \\
& ModelScope               & 8{,}858 & 3{,}100 & 0   & 11{,}958 \\
& Pika                     & 11{,}232 & 3{,}931 & 0 & 15{,}163 \\
& Text2Video-Zero          & 7{,}727 & 2{,}705 & 0   & 10{,}432 \\
& VideoCrafter2            & 24{,}215 & 8{,}475 & 0  & 32{,}690 \\
& CogVideoX1.5-5B-T        & 0 & 0 & 52  & 52 \\
& HunyuanVideo             & 0 & 0 & 76  & 76 \\
& HunyuanVideo-I2V         & 0 & 0 & 91  & 91 \\
& LTX-Video-13B-I          & 0 & 0 & 80  & 80 \\
& LTX-Video-13B-T          & 0 & 0 & 34  & 34 \\
& SkyReels-V2              & 0 & 0 & 59  & 59 \\
& SkyReels-V2-I2V-14B-540P & 0 & 0 & 83  & 83 \\
& Wan2.1-T2V-1.3B          & 0 & 0 & 76  & 76 \\
& Wan2.1-VACE-1.3B-T       & 0 & 0 & 86  & 86 \\
& Wan2.2-I2V-14B           & 0 & 0 & 86  & 86 \\
& Wan2.2-T2V-14B           & 0 & 0 & 86  & 86 \\
& Wan2.2-TI2V-5B-I         & 0 & 0 & 86  & 86 \\
& Wan2.2-TI2V-5B-T         & 0 & 0 & 83  & 83 \\
& Runway Gen-4 Turbo       & 0 & 0 & 49  & 49 \\
& Hailuo AI                & 0 & 0 & 59  & 59 \\
& Kling v1                 & 0 & 0 & 62  & 62 \\
& Pika 2.2                 & 0 & 0 & 62  & 62 \\
& PixVerse v4.5            & 0 & 0 & 64  & 64 \\
& Sora 2                   & 0 & 0 & 70  & 70 \\
& Seedance 2.0             & 0 & 0 & 576 & 576 \\

\midrule
\multicolumn{2}{@{}l}{\textbf{Total}}
& \textbf{68{,}906}
& \textbf{23{,}458}
& \textbf{7{,}636}
& \textbf{100{,}000} \\
\bottomrule
\end{tabular*}

\caption{Composition of the proposed \textbf{FaceVid-Forensics-100K} dataset
across the training, in-domain test, and out-of-distribution (OOD) splits, broken
down by source and synthesis method. The fake subset contains face-swapping (FS),
face-reenactment (FR), and entire-face synthesis (EFS) videos.}
\label{tab:dataset}
\end{table*}

\section{Additional Main Experimental Results}
\label{sec:additional-experimental-results}

\subsection{Detailed Out-of-Distribution Results}
\label{sec:additional-ood-results}

\begin{table*}[p]
\centering
\renewcommand{\arraystretch}{0.92}
\providecommand{\evalrot}[1]{\rotatebox{60}{#1}}

\begin{tabular*}{\textwidth}{@{\extracolsep{\fill}}l*{10}{c}@{}}
\toprule
\raisebox{2.5ex}{\textbf{Method}} & \evalrot{\textbf{CogX}} & \evalrot{\textbf{HunV}} & \evalrot{\textbf{HunI2V}} & \evalrot{\textbf{LTX-I}} & \evalrot{\textbf{LTX-T}} & \evalrot{\textbf{SkyV2}} & \evalrot{\textbf{SkyI2V}} & \evalrot{\textbf{W21T}} & \evalrot{\textbf{W21V}} & \evalrot{\textbf{W22}} \\
\midrule
\multicolumn{11}{@{}l}{\textbf{\textit{Small Vision Models}}} \\
TALL++ & \textbf{79.49} & 70.93 & 47.48 & 60.74 & 71.40 & \underline{73.34} & 52.23 & 68.30 & 62.92 & 50.71 \\
TFCU & 71.60 & 66.64 & 51.41 & 58.81 & 68.15 & 62.81 & 52.38 & 67.30 & 57.44 & 49.89 \\
DFD-FCG & 69.72 & 61.77 & 50.49 & 54.14 & 61.35 & 61.05 & 53.87 & 64.40 & 64.08 & 50.71 \\
DFGaze & 65.98 & 60.21 & 51.00 & 57.08 & 65.36 & 60.42 & 54.33 & 61.52 & 58.75 & 50.62 \\
Effort & 75.37 & 67.58 & 50.37 & 57.15 & 65.64 & 69.41 & 54.96 & \textbf{71.53} & 62.81 & 50.60 \\
\midrule
\multicolumn{11}{@{}l}{\textbf{\textit{Open-source MLLMs}}} \\
Qwen3.5-0.8B & 40.53 & 47.92 & 42.04 & 41.64 & 45.79 & 41.75 & 37.92 & 41.34 & 45.52 & 38.54 \\
Qwen3.6-35B-A3B & 51.22 & 53.80 & 49.57 & 46.89 & 50.20 & 44.24 & 48.02 & 51.17 & 48.49 & 46.16 \\
MiMo-V2.5 & 51.01 & 48.28 & 50.60 & 47.98 & 45.30 & 45.74 & 47.17 & 46.96 & 46.42 & 47.59 \\
\midrule
\multicolumn{11}{@{}l}{\textbf{\textit{Closed-source MLLMs}}} \\
GPT-5.5 & 47.96 & 48.72 & 46.31 & 45.99 & 44.11 & 47.50 & 47.13 & 46.74 & 47.60 & 46.44 \\
Gemini 3.1 Pro & 65.33 & 62.64 & 61.76 & 64.75 & 61.37 & 67.25 & 63.03 & 64.62 & 64.92 & \textbf{66.09} \\
GPT-4o & 58.60 & 57.18 & 58.33 & 59.32 & 53.62 & 57.79 & 54.92 & 59.82 & 57.21 & 58.96 \\
Gemini 3.5 Flash & 67.62 & 56.33 & \textbf{69.13} & 60.84 & 51.61 & 51.76 & 43.60 & \underline{70.81} & 63.62 & 57.22 \\
\midrule
\multicolumn{11}{@{}l}{\textbf{\textit{Forensics-tuned MLLMs}}} \\
Skyra & 63.75 & 65.21 & \underline{66.91} & \textbf{67.35} & 69.01 & 71.75 & 64.75 & 67.85 & 63.50 & 62.34 \\
VideoVeritas & 60.70 & 56.55 & 53.01 & 52.15 & 55.16 & 56.53 & 52.13 & 61.82 & 56.19 & 52.70 \\
\midrule
\multicolumn{11}{@{}l}{\textbf{\textit{Multi-Agent System}}} \\
Ours (w/o Video, SFT) & 67.68 & 67.23 & 57.65 & 61.86 & 68.08 & 63.17 & 64.06 & 64.59 & 66.68 & 56.79 \\
Ours (w/o Video, +GRPO) & \underline{79.14} & \textbf{73.73} & 61.28 & 65.54 & \textbf{74.62} & 72.20 & \textbf{67.44} & 70.44 & \underline{70.96} & 61.65 \\
Ours (w/ Video, SFT) & 69.30 & 69.20 & 62.71 & \underline{66.90} & 67.26 & 68.06 & 64.11 & 67.22 & 69.66 & 60.94 \\
Ours (w/ Video, +GRPO) & 74.15 & \underline{73.69} & 61.37 & 66.45 & \underline{71.60} & \textbf{73.87} & \underline{65.70} & 69.74 & \textbf{72.56} & \underline{62.67} \\
\bottomrule
\end{tabular*}

\vspace{0.3em}

\begin{tabular*}{\textwidth}{@{\extracolsep{\fill}}l*{10}{c}@{}}
\toprule
\raisebox{2.5ex}{\textbf{Method}} & \evalrot{\textbf{W22T}} & \evalrot{\textbf{W22I5}} & \evalrot{\textbf{W22T5}} & \evalrot{\textbf{Gen4}} & \evalrot{\textbf{Hailuo}} & \evalrot{\textbf{Kling}} & \evalrot{\textbf{Pika2}} & \evalrot{\textbf{PixV4.5}} & \evalrot{\textbf{Sora2}} & \evalrot{\textbf{Seed2.0}} \\
\midrule
\multicolumn{11}{@{}l}{\textbf{\textit{Small Vision Models}}} \\
TALL++ & \textbf{75.13} & 53.03 & 60.06 & 53.21 & 62.33 & 59.73 & \underline{81.50} & \textbf{79.49} & 58.42 & 64.04 \\
TFCU & 59.19 & 51.05 & 65.63 & 56.74 & 56.03 & 62.88 & \textbf{83.04} & 70.22 & 49.70 & \textbf{73.77} \\
DFD-FCG & 57.11 & 59.43 & 68.93 & 50.72 & 61.05 & 57.12 & 77.29 & 57.58 & 50.21 & 66.09 \\
DFGaze & 62.24 & 50.62 & 53.73 & 57.91 & 56.18 & 59.80 & 75.13 & 60.21 & 49.85 & 54.13 \\
Effort & 66.88 & 56.41 & \textbf{73.03} & 53.67 & 63.48 & 56.20 & 77.98 & 66.84 & 53.67 & 67.79 \\
\midrule
\multicolumn{11}{@{}l}{\textbf{\textit{Open-source MLLMs}}} \\
Qwen3.5-0.8B & 40.29 & 36.22 & 45.15 & 42.67 & 36.67 & 41.62 & 44.04 & 37.89 & 38.69 & 39.97 \\
Qwen3.6-35B-A3B & 46.74 & 44.42 & 51.03 & 52.63 & 44.24 & 45.88 & 46.69 & 50.02 & 39.57 & 65.55 \\
MiMo-V2.5 & 48.75 & 46.42 & 45.97 & 47.46 & 44.05 & 44.77 & 46.39 & 45.48 & 47.35 & 56.16 \\
\midrule
\multicolumn{11}{@{}l}{\textbf{\textit{Closed-source MLLMs}}} \\
GPT-5.5 & 45.86 & 46.44 & 47.13 & 45.13 & 47.50 & 45.73 & 44.92 & 44.89 & 45.54 & 49.23 \\
Gemini 3.1 Pro & 62.02 & 63.76 & 67.25 & 61.13 & 62.17 & 65.64 & 63.22 & 64.91 & 62.96 & 66.64 \\
GPT-4o & 51.40 & 56.05 & 56.72 & 61.94 & 51.85 & 56.80 & 51.96 & 54.64 & 55.30 & 60.28 \\
Gemini 3.5 Flash & 61.87 & \textbf{68.27} & \underline{70.71} & 54.37 & 59.39 & 63.80 & 70.26 & 60.53 & 61.11 & 67.30 \\
\midrule
\multicolumn{11}{@{}l}{\textbf{\textit{Forensics-tuned MLLMs}}} \\
Skyra & 66.99 & 60.01 & 61.14 & \textbf{69.97} & 68.36 & \textbf{72.09} & 59.19 & 66.57 & \textbf{69.76} & 48.17 \\
VideoVeritas & 56.77 & 48.63 & 57.55 & 50.03 & 50.60 & 54.69 & 57.91 & 62.15 & 52.68 & 64.49 \\
\midrule
\multicolumn{11}{@{}l}{\textbf{\textit{Multi-Agent System}}} \\
Ours (w/o Video, SFT) & 59.12 & 59.70 & 65.26 & 62.29 & 64.86 & 68.46 & 69.26 & 65.46 & 64.63 & 63.64 \\
Ours (w/o Video, +GRPO) & 62.24 & 65.14 & 67.44 & 60.42 & \underline{70.50} & \underline{71.20} & 76.85 & 70.85 & 67.77 & 65.29 \\
Ours (w/ Video, SFT) & 64.42 & \underline{66.75} & 69.53 & \underline{67.02} & 67.21 & 69.92 & 70.73 & 68.46 & 66.72 & 67.24 \\
Ours (w/ Video, +GRPO) & \underline{69.07} & 61.51 & 70.52 & 61.61 & \textbf{73.87} & 69.28 & 74.92 & \underline{75.05} & \underline{69.67} & \underline{72.10} \\
\bottomrule
\end{tabular*}
\caption{Detection accuracy on the 20 unseen video generators of the OOD test set, presented across two sub-tables (top: generators 1--10; bottom: generators 11--20). The best and second-best results are bolded and underlined, respectively.}
\label{tab:additional-ood}
\end{table*}

The per-generator accuracies in~\Cref{tab:additional-ood} indicate that the generalization benefit of the framework holds at the level of individual generators. After GRPO, our framework reaches a macro-average accuracy of 68.73\% when the judge agent receives only the outputs of the four observation agents, and 69.47\% when the sampled video frames are additionally provided. These two settings rank first and second on the macro average among all evaluated methods. With the sampled frames provided, our framework surpasses TALL++~\cite{tall++}, the strongest small vision model on average, by 5.25, and Skyra, the strongest forensics-tuned MLLM, by 4.24. Relative to the corresponding SFT settings, GRPO raises the macro average by 4.71 without video and 2.30 with video, whereas adding the sampled frames contributes 3.15 under SFT and 0.74 after GRPO. Both policy optimization and direct visual evidence therefore contribute to cross-generator generalization. With the sampled frames provided, our framework ranks first or second on 11 of the 20 held-out generators after GRPO, and the two GRPO settings together achieve the best result on six generators, which suggests that the improvement extends across several generator families rather than being driven by a single source.

\subsection{In-Domain Results}
\label{sec:additional-indomain-results}

\begin{table}[!htbp]
\centering
\begin{tabular*}{\linewidth}{@{\extracolsep{\fill}}lccc@{}}
\toprule
\textbf{Method} & \textbf{Acc} & \textbf{Recall} & \textbf{F1} \\
\midrule
\multicolumn{4}{@{}l}{\textbf{Small Vision Models}} \\
TALL++ & 96.06 & \underline{99.36} & 99.00 \\
TFCU & \textbf{98.66} & 99.31 & \textbf{99.46} \\
DFD-FCG & 97.86 & 99.07 & 99.22 \\
DFGaze & \underline{98.52} & 98.51 & 99.11 \\
Effort & 97.75 & \textbf{99.50} & \underline{99.37} \\
\midrule
\multicolumn{4}{@{}l}{\textbf{Open-source MLLMs}} \\
Qwen3.5-0.8B & 53.99 & 55.39 & 67.40 \\
Qwen3.6-35B-A3B & 74.84 & 83.33 & 87.85 \\
MiMo-V2.5 & 70.52 & 51.32 & 66.97 \\
\midrule
\multicolumn{4}{@{}l}{\textbf{Closed-source MLLMs}} \\
GPT-5.5 & 65.05 & 38.16 & 54.64 \\
Gemini 3.1 Pro & 61.57 & 94.58 & 90.90 \\
\midrule
\multicolumn{4}{@{}l}{\textbf{Forensics-tuned MLLMs}} \\
Skyra & 48.26 & 3.83 & 7.29 \\
VideoVeritas & 61.31 & 88.20 & 87.93 \\
\midrule
\multicolumn{4}{@{}l}{\textbf{Multi-Agent System}} \\
Ours (w/o Video, SFT) & 78.19 & 98.22 & 95.30 \\
Ours (w/o Video, +GRPO) & 87.34 & 96.87 & 96.35 \\
Ours (w/ Video, SFT) & 79.08 & 98.07 & 95.39 \\
Ours (w/ Video, +GRPO) & 84.31 & 97.78 & 96.19 \\
\bottomrule
\end{tabular*}
\caption{Overall performance on the in-domain test set. The best and second-best results are highlighted in bold and underlined.}
\label{tab:additional-indomain}
\end{table}

Specialized small vision models remain the strongest detectors on seen distributions in~\Cref{tab:additional-indomain}, with TFCU~\cite{tfcu} reaching 98.66\% Acc and 99.46\% F1. Among the MLLM-based methods, however, our framework performs best. After GRPO, it reaches 87.34\% Acc and 96.35\% F1 when the judge receives only the outputs of four observation agents, and 84.31\% Acc and 96.19\% F1 when the sampled video frames are additionally provided to the judge. Relative to the corresponding SFT settings, GRPO raises Acc by 9.15 and 5.23 and F1 by 1.05 and 0.80 in the two settings, respectively, while recall remains above 96.8\%. Staged Observation--Judge training and policy optimization yield competitive performance on known manipulations, while our framework generalizes better to unseen generators and remains more interpretable than small vision baselines.

\subsection{Qualitative Comparison}
\label{sec:qualitative-comparison}

\begin{figure*}[!t]
    \centering
    \includegraphics[width=\textwidth]{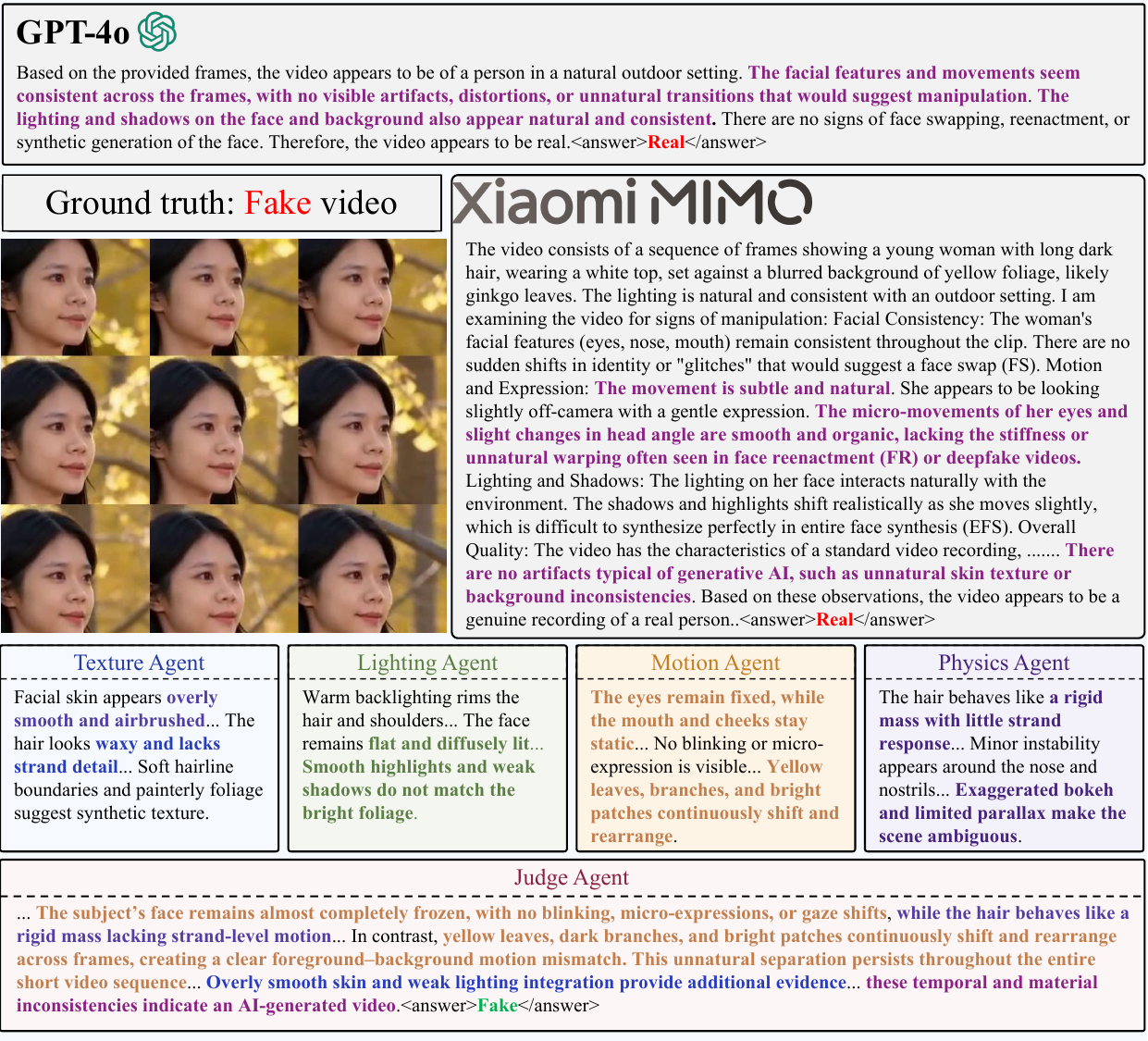}
    \caption{Qualitative comparison on a challenging fake video. GPT-4o and
    MiMo-V2.5 incorrectly predict Real, treating the stable facial appearance
    and natural-looking scene as evidence of authenticity. In contrast, the four
    observation agents identify complementary texture, lighting, motion, and
    physical-plausibility cues, and the judge agent reconciles their
    observations, emphasizes the persistent foreground--background motion
    mismatch, and correctly predicts Fake. Colored text highlights the evidence
    emphasized by each model.}
    \label{fig:qualitative-comparison}
\end{figure*}

As shown in~\Cref{fig:qualitative-comparison}, the fake video contains no obvious face-swapping boundary or severe frame-level distortion. GPT-4o~\cite{gpt4o} and MiMo-V2.5~\cite{mimov25} therefore rely primarily on the stable facial appearance and the natural-looking outdoor illumination, interpreting the absence of conspicuous artifacts as evidence of authenticity. Both models overlook the inconsistencies that become apparent only when the subject and the scene are compared across frames.

The four observation agents recover these weak but complementary cues. The texture observation agent identifies over-smoothed skin, waxy hair, soft hairline boundaries, and painterly foliage. The lighting observation agent notes that the flat facial illumination and the weak shadows are not fully integrated with the bright backlit background. The motion observation agent observes an almost fixed gaze and expression despite continuous changes in the leaves, branches, and bright background regions. The physics observation agent further reports rigid hair dynamics and limited parallax. Rather than relying on any single artifact, the judge agent reconciles these observations and uses the persistent foreground--background motion mismatch as the principal cue, with the texture, lighting, and physical anomalies providing corroborating evidence. This example illustrates how independent evidence collection and explicit reconciliation can turn individually subtle cues into a coherent and correct forensic decision.

\subsection{Evaluation of Textual Explanation Quality}
\label{sec:explanation-quality}

Following the VidGuard-R1~\cite{VidGuard-R1}, we evaluate explanation quality on a fixed subset of the OOD test set, obtained by randomly sampling 100 real and 100 fake videos. For each detector output, two independent judges---the open-source DeepSeek-V4 Pro~\cite{deepseek-v4} and the closed-source GPT-5-mini~\cite{openai2025gpt5}---score the generated rationale, so that the assessment does not rest on a single model family or provider. Both judges receive the same ground-truth answer, reference rationale, model answer, and model rationale, and both apply an identical rubric covering evidence accuracy, reference alignment, specificity and grounding, clarity and conciseness, and consistency between the rationale and the verdict. Each judge returns a single holistic integer score from 1 to 10. \Cref{tab:explanation-quality} reports the mean scores for real videos, fake videos, and the complete 200-video subset. The shared evaluator prompt is provided in Listing~\ref{lst:explanation-quality-evaluator-prompt}.

\begin{table*}[!htbp]
\centering
\footnotesize
\renewcommand{\arraystretch}{1.02}
\begin{tabular*}{\textwidth}{@{\extracolsep{\fill}}lccc ccc@{}}
\toprule
\multirow{2}{*}[-0.5em]{\textbf{Method}}
& \multicolumn{3}{c}{\textbf{DeepSeek-V4 Pro}}
& \multicolumn{3}{c}{\textbf{GPT-5-mini}} \\
\cmidrule(lr){2-4}\cmidrule(lr){5-7}
& \textbf{Fake} & \textbf{Real} & \textbf{Overall}
& \textbf{Fake} & \textbf{Real} & \textbf{Overall} \\
\midrule
\multicolumn{7}{@{}l}{\textbf{Open-source MLLMs}} \\
Qwen3.6-35B-A3B & 3.49 & 5.30 & 4.39 & 5.02 & 6.19 & 5.61 \\
MiMo-V2.5 & 2.00 & \textbf{6.84} & 4.42 & 3.76 & \underline{7.30} & 5.53 \\
\midrule
\multicolumn{7}{@{}l}{\textbf{Closed-source MLLMs}} \\
GPT-4o & 2.65 & 4.56 & 3.60 & 4.37 & 6.07 & 5.22 \\
Gemini 3.1 Pro & 7.15 & 3.65 & 5.40 & \underline{7.45} & 5.02 & 6.24 \\
Gemini 3.5 Flash & 5.98 & 6.52 & \underline{6.25} & 6.64 & 7.12 & \textbf{6.88} \\
\midrule
\multicolumn{7}{@{}l}{\textbf{Forensics-tuned MLLMs}} \\
Skyra & 3.16 & \underline{6.58} & 4.87 & 4.70 & \textbf{7.35} & 6.03 \\
VideoVeritas & \textbf{8.05} & 4.07 & 6.06 & \textbf{7.90} & 5.31 & \underline{6.61} \\
\midrule
\multicolumn{7}{@{}l}{\textbf{Multi-Agent System}} \\
Ours (w/o Video, SFT) & 7.66 & 4.58 & 6.12 & 6.99 & 5.12 & 6.05 \\
Ours (w/o Video, SFT+GRPO) & 6.40 & 4.46 & 5.43 & 6.67 & 4.85 & 5.76 \\
Ours (w/ Video, SFT) & \underline{7.89} & 4.84 & \textbf{6.37} & 7.27 & 5.22 & 6.25 \\
Ours (w/ Video, SFT+GRPO) & 7.75 & 4.14 & 5.95 & 7.37 & 4.68 & 6.03 \\
\bottomrule
\end{tabular*}
\caption{Mean explanation-quality scores on a subset of the OOD test set
containing 100 randomly selected real and 100 randomly selected fake videos. The best and second-best results in each column are bolded and underlined.}
\label{tab:explanation-quality}
\end{table*}

DeepSeek-V4 Pro assigns the highest overall score (6.37) to our SFT configuration with video input, while GPT-5-mini assigns it 6.25. Within our framework, providing video input improves the overall score both with and without GRPO optimization across both judges, suggesting better grounding of generated rationales in visual evidence. GRPO-optimized configurations remain competitive, although GRPO primarily targets classification performance rather than the explanation-quality rubric; therefore, these scores should be interpreted together with detection performance. All four configurations receive higher scores on fake than on real videos, but this pattern is specific to the sampled subset and should not be generalized more broadly.

\section{Reasoning Strategies}
\label{sec:reasoning-strategies}

\begin{figure}
    \centering
    \includegraphics[scale=0.8]{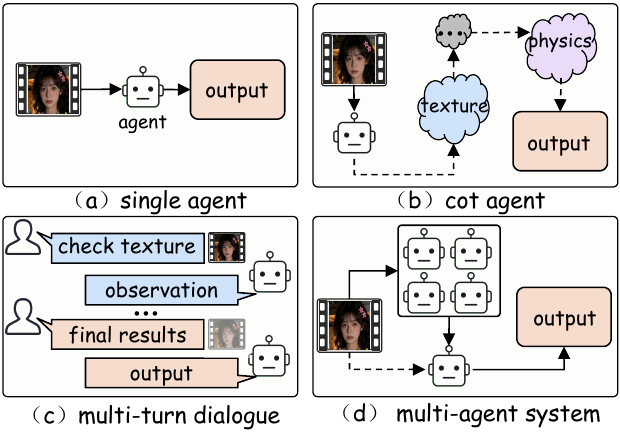}
    \caption{Comparison of the reasoning strategies evaluated in our experiments: (a) single-agent direct prediction, (b) single-agent chain-of-thought reasoning, (c) multi-turn dialogue, and (d) our multi-agent system with four independent observation agents and a separate judge agent.}
    \label{fig:reasoning-strategy-overview}
\end{figure}

\Cref{fig:reasoning-strategy-overview} summarizes the four reasoning structures evaluated in our experiments. Single directly predicts the video label from the sampled frames. CoT uses a single agent and a single interaction, but explicitly prompts the model to reason step by step over the four forensic dimensions before giving the final verdict. Multi-turn dialogue instead queries the same agent through multiple turns: dimension-specific observations are collected in earlier turns and then synthesized into a final prediction in a subsequent turn. In contrast, our multi-agent system assigns the four forensic dimensions to independent observation agents and uses a separate judge agent to reconcile their evidence and issue a final verdict.

\paragraph{Single.}
The Single baseline maps the sampled frames directly to an explanation and a prediction in a single model call. The general-purpose open- and closed-source MLLMs use the shared prompts in Listing~\ref{lst:general-mllm-prompt}, with the sampled frames and their timestamps appended to the user message.

\paragraph{Chain-of-Thought.}
The CoT baseline performs the four forensic analyses and produces the final decision in a single model call. Its exact system and user prompts are shown in Listing~\ref{lst:cot-prompt}.

\paragraph{Multi-turn.}
The Multi-turn baselines use a shared system prompt and five user turns: the first four turns examine texture, lighting, motion, and physical plausibility, respectively, while the fifth turn produces the final explanation and verdict. The three variants differ only in when video frames are provided to the model, as summarized in~\Cref{tab:multiturn-configs}. In \textit{Multi-turn}, frames are provided only in the first turn, and all subsequent turns rely on the retained dialogue context and text-only prompts. In \textit{Multi-turn-Obs}, frames are provided during the four observation turns, allowing each aspect-specific analysis to directly access the visual evidence, whereas the final verdict turn is text-only and aggregates the preceding observations. In \textit{Multi-turn-All}, frames are provided in all five turns, including the final verdict turn, so the model can revisit the visual evidence when generating the explanation and prediction. All three settings use the same first four prompts in Listing~\ref{lst:multiturn-shared-prompt} and differ only in the final prompt, as shown in Listing~\ref{lst:multiturn-verdict-prompts}.

\noindent\begin{minipage}{\columnwidth}
\centering
\begin{tabular*}{\columnwidth}{@{\extracolsep{\fill}}lcc@{}}
\toprule
\textbf{Setting} & \makecell[c]{\textbf{Turns with Frames}}
& \makecell{\textbf{Turn-5 Prompt}} \\
\midrule
Multi-turn      & 1    & Text-only \\
Multi-turn-Obs  & 1--4 & Text-only \\
Multi-turn-All  & 1--5 & Video-aware \\
\bottomrule
\end{tabular*}
\captionof{table}{Frame-delivery settings for Multi-turn inference. Turns not listed in the middle column receive text only, and the preceding textual responses remain in the dialogue history in every setting.}
\label{tab:multiturn-configs}
\end{minipage}

\subsection{Efficiency Analysis}

We evaluate latency and throughput on 100 randomly sampled OOD videos (50 real and 50 fake) using a single NVIDIA RTX 5090 GPU. As shown in~\Cref{tab:inference-efficiency}, Single and CoT have the lowest latency, while Multi-turn has the highest latency on both MLLMs. The latency of our two configurations lies between these baselines, and both are faster than Multi-turn. This difference is particularly clear on InternVL3.5-8B, where our latency is 4.781--5.486 seconds per video, compared with 14.150 seconds for Multi-turn. Our framework also achieves 151.293--181.88 tokens per second across the two MLLMs, approximately twice the throughput of Multi-turn. Thus, although our framework is slower than the one-pass strategies, parallel execution of the observation agents keeps its additional inference cost moderate. The configuration without video input further reduces latency by avoiding video processing in the judge agent. Furthermore, since our multi-agent framework introduces more prompts compared to single-agent methods, future work could consider applying prompt compression techniques such as BEAVER~\cite{hu2026beavertrainingfreehierarchicalprompt} to further optimize inference efficiency.

\begin{table}[!htbp]
\centering
\begin{tabular*}{\linewidth}{@{\extracolsep{\fill}}lrr@{}}
\toprule
\textbf{Strategy}
& \makecell[c]{\textbf{Latency} $\downarrow$\\\textbf{(s/video)}}
& \makecell[c]{\textbf{Throughput} $\uparrow$\\\textbf{(tokens/s)}} \\
\midrule
\multicolumn{3}{@{}l}{\textbf{Qwen2.5-VL-7B}} \\
Single & 1.503 & 79.10 \\
CoT & 2.095 & 85.27 \\
Multi-turn & 7.123 & 90.47 \\
Ours (w/ Video) & 6.128 & 172.26 \\
Ours (w/o Video) & 5.791 & 181.88 \\
\midrule
\multicolumn{3}{@{}l}{\textbf{InternVL3.5-8B}} \\
Single & 1.583 & 53.892 \\
CoT & 2.864 & 70.869 \\
Multi-turn & 14.150 & 84.581 \\
Ours (w/ Video) & 5.486 & 151.293 \\
Ours (w/o Video) & 4.781 & 168.329 \\
\bottomrule
\end{tabular*}
\caption{Efficiency comparison of different reasoning strategies on an NVIDIA
RTX 5090 GPU.}
\label{tab:inference-efficiency}
\end{table}

\subsection{Qualitative Comparison}
\label{sec:reasoning-strategy-qualitative}

\begin{figure*}[t]
    \centering
    \includegraphics[width=\textwidth]{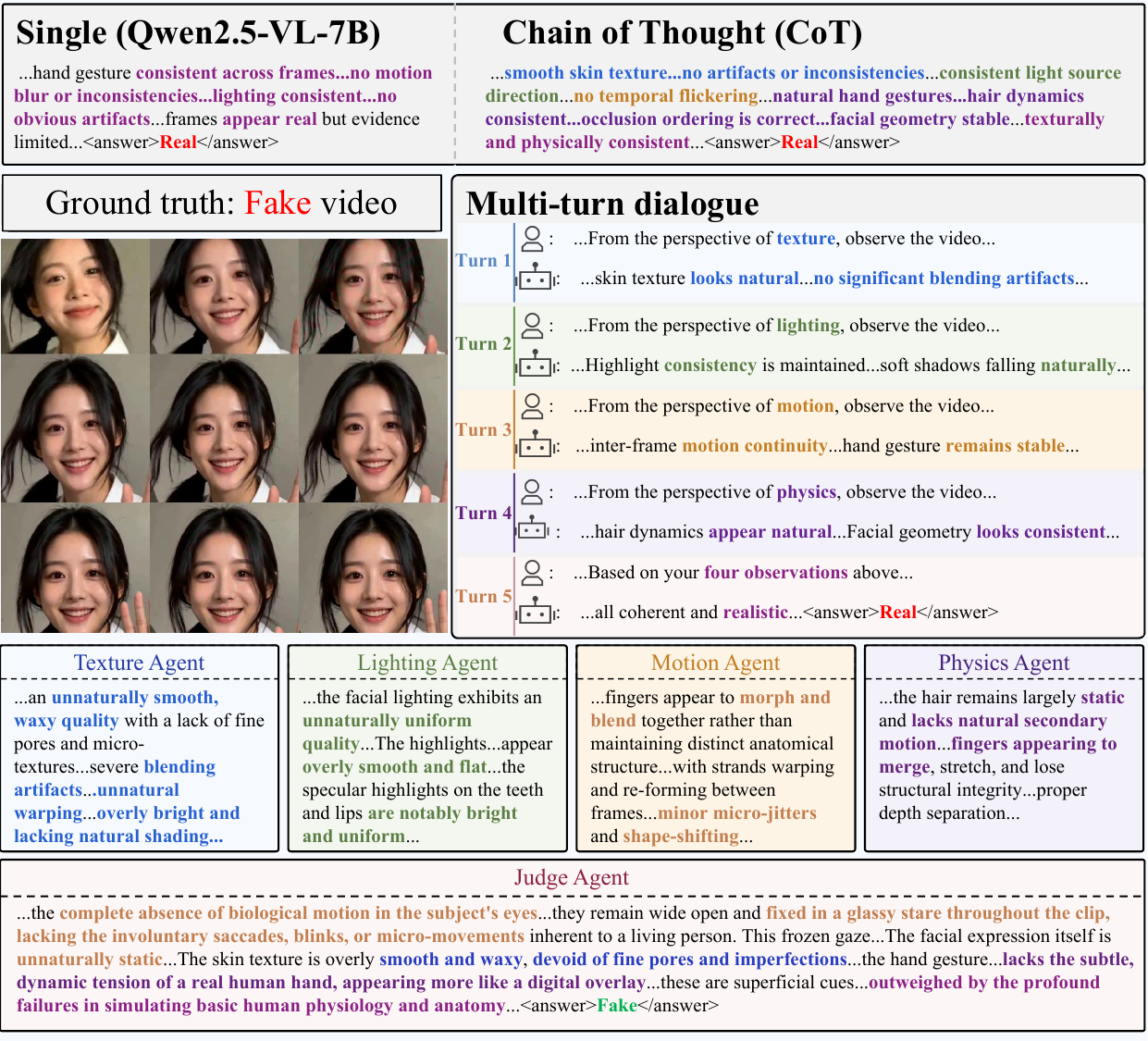}
    \caption{Qualitative comparison of Single prediction, CoT, Multi-turn dialogue, and our
    multi-agent system (MAS) on a fake video. Single, CoT, and Multi-turn cite plausible
    but superficial evidence of authenticity and incorrectly predict Real. In contrast, the
    independent observation agents identify complementary anomalies in texture,
    lighting, motion, and physical plausibility, which the judge agent
    reconciles to correctly predict Fake.}
    \label{fig:reasoning-strategy-qualitative}
\end{figure*}

As shown in~\Cref{fig:reasoning-strategy-qualitative}, Single prediction, CoT,
and Multi-turn dialogue all interpret the smooth facial appearance, stable geometry,
coherent illumination, and apparently natural motion as evidence that the video is real.
Specifically, Single prediction relies on a coarse overview and prematurely declares the video
authentic despite acknowledging limited evidence. Step-by-step reasoning via CoT or
decomposing the analysis into successive dialogue turns fails to correct this initial assessment:
the single-agent trajectory repeatedly reinforces benign interpretations across all forensic
dimensions, ultimately repeating the incorrect Real verdict.

Our MAS instead elicits complementary evidence from independent observation
agents. The texture agent identifies waxy skin, missing
micro-texture, and blending artifacts; the lighting agent reports
unnaturally uniform facial illumination; the motion agent detects
morphing fingers and temporal shape-shifting; and the physics agent
highlights static hair and implausible hand dynamics. The judge agent reconciles
these observations and further emphasizes the frozen gaze and the absence of
natural biological micro-motion, which yields the correct Fake prediction. This
example illustrates that the advantage of the MAS comes not merely from
performing more reasoning steps, but from collecting diverse forensic evidence
independently and reconciling it explicitly before prediction.

\section{Observation and Judge Model Combinations}
\label{sec:observation-judge-models}

\begin{table}[!htbp]
\centering
\renewcommand{\arraystretch}{1.08}
\begin{tabular*}{\linewidth}{@{\extracolsep{\fill}}llccc@{}}
\toprule
\textbf{Observation} & \textbf{Judge} & \textbf{Acc} & \textbf{Recall} & \textbf{F1} \\
\midrule
\multicolumn{5}{@{}l@{}}{\textbf{w/o Video}} \\
InternVL3.5 & InternVL3.5
& 62.72
& \underline{85.52}
& 46.94 \\
InternVL3.5 & Qwen2.5-VL
& \underline{63.50}
& 85.15
& \textbf{47.54} \\
Qwen2.5-VL & InternVL3.5
& 60.70
& \textbf{86.93}
& 45.51 \\
Qwen2.5-VL & Qwen2.5-VL
& \textbf{63.61}
& 82.24
& \underline{47.53} \\
\midrule
\multicolumn{5}{@{}l@{}}{\textbf{w/ Video}} \\
InternVL3.5 & InternVL3.5
& \textbf{69.07}
& \underline{83.80}
& \textbf{52.43} \\
InternVL3.5 & Qwen2.5-VL
& 67.95
& 78.33
& 51.43 \\
Qwen2.5-VL & InternVL3.5
& \underline{68.89}
& 82.45
& \underline{52.28} \\
Qwen2.5-VL & Qwen2.5-VL
& 67.03
& \textbf{87.76}
& 50.49 \\
\bottomrule
\end{tabular*}
\caption{Performance of the Observation--Judge model combinations on the OOD test
set. InternVL3.5 is InternVL3.5-8B and Qwen2.5-VL is Qwen2.5-VL-7B. All
configurations use SFT only.}
\label{tab:observation-judge-models}
\end{table}

We further evaluate whether the observation agents and the judge agent need to come from the same MLLM family. As shown in~\Cref{tab:observation-judge-models}, heterogeneous combinations remain competitive with homogeneous ones. When the judge agent receives only the observation agent outputs, InternVL3.5~\cite{internvl} observations judged by Qwen2.5-VL~\cite{qwen25vl} achieve the best F1 of 47.54\%. When the sampled video frames are additionally provided, InternVL3.5 used for both components performs best, reaching 69.07\% Acc and 52.43\% F1, while the heterogeneous Qwen2.5-VL--InternVL3.5 combination achieves comparable results. Providing the sampled frames to the judge agent improves Acc and F1 for every combination, indicating that the framework is modular with respect to the MLLM used for each component.

\section{Impact of Model Parameter Scale}
\label{sec:model-scale-ablation}

We evaluate the scalability of our framework across different parameter scales using the 3B, 7B, and 32B variants of Qwen2.5-VL. To isolate the effect of model capacity, all variants are trained under the SFT protocol across both Judge Agent settings, in which the Judge Agent receives either only the observation reports or these reports together with the sampled video frames.

As shown in \Cref{tab:model-scale-ablation}, scaling the base MLLM from 3B to 32B parameters yields consistent gains in accuracy and F1 score. When the Judge Agent receives only the observation reports, accuracy increases from 61.36\% (3B) to 63.61\% (7B) and reaches 66.40\% (32B), while F1 score steadily improves from 45.81\% to 49.92\%. A similar scaling trend is observed when the sampled video frames are additionally provided, with F1 score reaching a peak of 50.81\% at 32B. Notably, as model capacity grows, the text-only Judge Agent becomes increasingly adept at reconciling textual forensic reports, narrowing the gap with the video-aware setting while maintaining solid interpretability.

\begin{table}[!htbp]
\centering
\setlength{\tabcolsep}{1.8pt}
\begin{tabular*}{\linewidth}{@{\extracolsep{\fill}}l ccc ccc@{}}
\toprule
 & \multicolumn{3}{c}{\textbf{w/o Video}} & \multicolumn{3}{c}{\textbf{w/ Video}} \\
\cmidrule(lr){2-4} \cmidrule(lr){5-7}
\multirow{-2}{*}[0.6ex]{\textbf{Parameter}} & \textbf{Acc} & \textbf{Recall} & \textbf{F1} & \textbf{Acc} & \textbf{Recall} & \textbf{F1} \\
\midrule
3B  & 61.36 & \textbf{83.33} & 45.81 & 66.65 & \underline{82.55} & 50.16 \\
7B  & \underline{63.61} & \underline{82.24} & \underline{47.53} & \underline{67.03} & \textbf{87.76} & \underline{50.49} \\
32B & \textbf{66.40} & 80.36 & \textbf{49.92} & \textbf{67.36} & 81.47 & \textbf{50.81} \\
\bottomrule
\end{tabular*}
\caption{Performance comparison across different Qwen2.5-VL parameter scales on the OOD test set under SFT. The best and second-best results within each Judge Agent setting are highlighted in bold and underlined.}
\label{tab:model-scale-ablation}
\end{table}

\section{Prompt Templates}\label{sec:prompt-templates}

\noindent\begin{minipage}{\columnwidth}
\centering
\fontsize{8pt}{8.5pt}\selectfont
\begin{tabular*}{\columnwidth}{@{\extracolsep{\fill}}l >{\raggedright\arraybackslash}p{0.70\linewidth}@{}}
\toprule
\textbf{Agent} & \textbf{Prompt-specific fields} \\
\midrule
Texture
& \texttt{expert/perspective}: texture and detail; \texttt{focus}: texture;
\texttt{cues}: skin texture, edge sharpness, blending artifacts, material
consistency, and fine-grained detail stability across frames \\
Lighting
& \texttt{expert/perspective}: lighting; \texttt{focus}: lighting;
\texttt{cues}: light source direction, highlight consistency, shadow
placement, specular reflections, and overall illumination coherence \\
Motion
& \texttt{expert/perspective}: motion; \texttt{focus}: motion;
\texttt{cues}: inter-frame motion continuity, unnatural movements, temporal
flickering, or physically implausible actions \\
Physics
& \texttt{expert/perspective}: physical plausibility; \texttt{focus}: physics;
\texttt{cues}: hair dynamics, clothing behavior, occlusion ordering,
perspective correctness, and geometric deformation \\
\bottomrule
\end{tabular*}
\captionof{table}{Role-specific fields used to instantiate the shared Observation
Agent prompt templates of Listing~\ref{lst:observer-prompt-template} for each of
the four forensic dimensions.}
\label{tab:observer-prompt-fields}
\end{minipage}

\paragraph{Multi-Agent System.}
Each observation agent receives the sampled video frames and a prompt tailored
to one forensic dimension. The four prompts share the templates in
Listing~\ref{lst:observer-prompt-template}, and \Cref{tab:observer-prompt-fields}
provides the exact role-specific fields. The judge agent then receives the four
observations through a shared user template. Without video input, the judge agent
receives only these observations; with video input, it additionally receives the
sampled frames. The corresponding system prompts are shown in
Listing~\ref{lst:judge-prompt-templates}.

\paragraph{Observation Label Generation.}
For textual label construction, the annotator MLLMs use the same
dimension-specific prompts given in Listing~\ref{lst:observer-prompt-template}
and \Cref{tab:observer-prompt-fields}. Their user messages additionally
contain the uniformly sampled frames and the corresponding timestamps. For each
dimension, DeepSeek-V4 Pro then combines the available
model reports using the prompts in
Listing~\ref{lst:observation-aggregation-prompt}. The ground-truth label is
provided only as guidance for resolving conflicting reports, and the output is
required to contain neither a verdict nor a reference to that label.

\paragraph{Explanation Evaluation.}
DeepSeek-V4 Pro and GPT-5-mini independently
evaluate the generated rationales using the same inputs, rubric, and output format. Each judge
receives the ground-truth answer, a reference rationale, the model answer, and
the model rationale, and assigns a holistic integer score from 1 to 10 based
on evidence accuracy, reference alignment, specificity and grounding, clarity
and conciseness, and verdict consistency. The shared evaluator prompt is shown
in Listing~\ref{lst:explanation-quality-evaluator-prompt}, and the sampling
protocol and results are reported in Section~\ref{sec:explanation-quality}.

\begin{promptsingle}{Prompts used by the Single baseline for the general-purpose
open- and closed-source MLLMs.}{lst:general-mllm-prompt}
\textbf{System Prompt}
\begin{lstlisting}[style=prompt]
You are an expert video analyst.
Please think about the question as if you were a human pondering deeply. It's encouraged to include self-reflection or verification in the reasoning process. Put the explanation of your judgment within <explanation></explanation> tags. Finally, give the final verdict within <answer></answer> tags.
\end{lstlisting}

\par\smallskip
\textbf{User Prompt}
\begin{lstlisting}[style=prompt,belowskip=0pt]
Is this video real or fake?

The following images are uniformly sampled frames from the video.
\end{lstlisting}
\end{promptsingle}

\begin{promptsingle}{Prompts used by the chain-of-thought (CoT) baseline.}
{lst:cot-prompt}
\textbf{System Prompt}
\begin{lstlisting}[style=prompt]
You are an expert face video forensics analyst. You are shown frames uniformly sampled from a video, in order. Determine whether the face video is Real or Fake based ONLY on the visible evidence in these frames.

Think step by step. Work through the following four analysis steps in order. In each step, report concrete, specific visual cues - mention frame ranges, facial regions, or objects where you see potential manipulation artifacts, or where everything appears consistent and natural. Do NOT give a real/fake verdict inside these four steps.

1. Texture and detail: skin texture, edge sharpness, blending artifacts around the face boundary, material consistency, and fine-grained detail stability across frames.
2. Lighting: light source direction, highlight consistency, shadow placement, specular reflections on skin and eyes, and overall illumination coherence between the face and the scene.
3. Motion: inter-frame motion continuity, unnatural movement, temporal flickering or jitter, and physically implausible actions across frames.
4. Physical plausibility: hair dynamics, clothing behavior, occlusion ordering, perspective correctness, facial geometry, and geometric deformation.

After the four steps, weigh the four observations together and reach a final decision. If clear manipulation artifacts appear in one or more steps, answer fake. If the face is texturally, temporally, and physically consistent with the scene, answer real.
\end{lstlisting}

\par\smallskip
\textbf{System Prompt (continued)}
\begin{lstlisting}[style=prompt]
Output your response strictly in the following format, with every tag present and non-empty:

<texture>your texture and detail observations</texture>
<lighting>your lighting observations</lighting>
<motion>your motion observations</motion>
<physics>your physical plausibility observations</physics>
<explanation>a brief rationale that synthesizes the four observations above into your decision</explanation>
<answer>real</answer>

Output requirements:
* All six tags are required and must not be empty.
* Each of <texture>, <lighting>, <motion>, <physics> must contain at least one concrete visual observation and must NOT contain a real/fake judgment.
\end{lstlisting}

\par\smallskip
\textbf{User Prompt}
\begin{lstlisting}[style=prompt,belowskip=0pt]
Analyze the provided video frames step by step. First examine:
1. Texture and detail (skin texture, edge sharpness, blending artifacts around the face boundary, material consistency, fine detail stability across frames).
2. Lighting (light source direction, highlight consistency, shadow placement, specular reflections on skin and eyes, illumination coherence).
3. Motion (inter-frame motion continuity, unnatural movements, temporal flickering or jitter, physically implausible actions across frames).
4. Physical plausibility (hair dynamics, clothing behavior, occlusion ordering, perspective correctness, facial geometry, geometric deformation).

First put your complete step-by-step rationale within <explanation></explanation> tags; do not put any answer or label inside the explanation. Then output your final verdict as either <answer>real</answer> or <answer>fake</answer>.
\end{lstlisting}
\end{promptsingle}

\begin{promptsingle}{System prompt and four observation-turn prompts shared by all three Multi-turn frame-delivery settings.}{lst:multiturn-shared-prompt}
\textbf{System Prompt}
\begin{lstlisting}[style=prompt]
You are an expert face video forensics analyst. Your task is to determine whether the face video is real or fake based only on the visible evidence in the provided frames.
\end{lstlisting}

\textbf{Turn 1: Texture and Detail}
\begin{lstlisting}[style=prompt]
These are frames uniformly sampled from the video, in order. From the perspective of texture and detail, observe the video and report any anomalies related to skin texture, edge sharpness, blending artifacts around the face boundary, material consistency, and fine-grained detail stability across frames. Be concrete - mention frame ranges, facial regions, or objects. Put your report inside <observation></observation> tags. Do not make a real/fake judgment yet.
\end{lstlisting}

\textbf{Turn 2: Lighting}
\begin{lstlisting}[style=prompt]
Now from the perspective of lighting, report any anomalies related to light source direction, highlight consistency, shadow placement, specular reflections on skin and eyes, and overall illumination coherence between the face and the scene. Be concrete - mention frame ranges, facial regions, or objects. Put your report inside <observation></observation> tags. Do not make a real/fake judgment yet.
\end{lstlisting}

\par\smallskip
\textbf{Turn 3: Motion}
\begin{lstlisting}[style=prompt]
Now from the perspective of motion, report any anomalies related to inter-frame motion continuity, unnatural movements, temporal flickering or jitter, and physically implausible actions across frames. Be concrete - mention frame ranges, facial regions, or objects. Put your report inside <observation></observation> tags. Do not make a real/fake judgment yet.
\end{lstlisting}

\textbf{Turn 4: Physical Plausibility}
\begin{lstlisting}[style=prompt,belowskip=0pt]
Now from the perspective of physical plausibility, report any anomalies related to hair dynamics, clothing behavior, occlusion ordering, perspective correctness, facial geometry, and geometric deformation. Be concrete - mention frame ranges, facial regions, or objects. Put your report inside <observation></observation> tags. Do not make a real/fake judgment yet.
\end{lstlisting}
\end{promptsingle}

\begin{promptsingle}{Final-turn prompts for the three Multi-turn frame-delivery settings.}{lst:multiturn-verdict-prompts}
\textbf{Text-only Verdict (Multi-turn, Multi-turn-Obs)}
\begin{lstlisting}[style=prompt,belowskip=0pt]
Based on your four observations above, decide whether the face video is real or fake. First put a brief explanation that synthesizes your four observations within <explanation></explanation> tags; do not put any answer or label inside the explanation. Then output exactly either <answer>real</answer> or <answer>fake</answer>, with no other text inside the answer tag.
\end{lstlisting}

\par\smallskip
\textbf{Image-aware Verdict (Multi-turn-All)}
\begin{lstlisting}[style=prompt,belowskip=0pt]
Based on your four observations above and the video frames, decide whether the face video is real or fake. First put a brief explanation that synthesizes your four observations within <explanation></explanation> tags; do not put any answer or label inside the explanation. Then output exactly either <answer>real</answer> or <answer>fake</answer>, with no other text inside the answer tag.
\end{lstlisting}
\end{promptsingle}

\begin{promptsingle}{Prompt templates shared by the four observation agents. The fields in braces are instantiated with the role-specific values listed in \Cref{tab:observer-prompt-fields}.}{lst:observer-prompt-template}
\textbf{Observation agent system prompt}
\begin{lstlisting}[style=prompt]
You are a {expert} analysis expert in a video forensics team. Your task is to carefully observe the video and report any anomalies related to {cues}. Focus ONLY on {focus}-related observations. Do NOT make a final real/fake judgment. Return only a concise report inside <observation></observation> tags.
\end{lstlisting}

\par\smallskip
\textbf{Observation agent user prompt}
\begin{lstlisting}[style=prompt,belowskip=0pt]
Observe the provided video from the perspective of {perspective} analysis. List the specific visual cues you noticed. Be concrete and precise - mention frame ranges, regions, or objects where you see potential issues or where everything appears normal. Put your report inside <observation></observation> tags. Do not make a final real/fake judgment and do not use <answer> tags. Keep your response concise (within 200 words).
\end{lstlisting}
\end{promptsingle}

\begin{promptsingle}{Judge agent prompts with and without video input. Both settings use the same user prompt, populated with the four observation agent reports; the sampled frames are additionally attached when video input is provided.}{lst:judge-prompt-templates}
\textbf{Judge System Prompt: w/o Video}
\begin{lstlisting}[style=prompt,belowskip=0pt]
You are the final judge for binary video forgery classification. You cannot access any images or video. Use only the four expert analysis reports supplied as text. Put a brief explanation of your judgment within <explanation></explanation> tags; do not put any <answer> tags or final label inside the explanation. Then output exactly one tagged lowercase label: <answer>real</answer> or <answer>fake</answer>.
\end{lstlisting}

\textbf{Judge System Prompt: w/ Video}
\begin{lstlisting}[style=prompt,belowskip=0pt]
You are the final judge for binary video forgery classification. You are given the video frames together with four expert analysis reports (texture, lighting, motion, physical plausibility) supplied as text. Weigh both the visual evidence and the four reports. Put a brief explanation of your judgment within <explanation></explanation> tags; do not put any <answer> tags or final label inside the explanation. Then output exactly one tagged lowercase label: <answer>real</answer> or <answer>fake</answer>.
\end{lstlisting}

\par\smallskip
\textbf{Shared Judge User Prompt}
\begin{lstlisting}[style=prompt,belowskip=0pt]
Texture analysis report:
{texture_report}

Lighting analysis report:
{lighting_report}

Motion analysis report:
{motion_report}

Physical plausibility analysis report:
{physics_report}

Return a brief explanation in <explanation></explanation> tags. Do not put any <answer> tags or final label inside the explanation. Then output exactly <answer>real</answer> or <answer>fake</answer>.
\end{lstlisting}
\end{promptsingle}

\begin{promptsingle}{DeepSeek-V4 Pro prompts for aggregating the annotator reports of one forensic dimension into a single observation label. The aggregation is performed independently for texture, lighting, motion, and physics.}{lst:observation-aggregation-prompt}
\textbf{DeepSeek-V4 Pro System Prompt}
\begin{lstlisting}[style=prompt]
You are a video forensics expert. Multiple AI models have each analyzed the same video from a specific analytical perspective and produced independent observation reports. Your task is to read all of their reports and synthesize them into a single, coherent, comprehensive observation.

Aggregate multi-agent observations with the video label (real or fake) in mind: use the ground-truth label solely as internal guidance to resolve conflicts between models when their reports disagree.

Rules:
- Write ONE unified observation paragraph (100-200 words).
- Do NOT list models by name (e.g. "Model 1 said..."). Present the observations as a single cohesive expert analysis.
- Resolve conflicts: when models disagree, prioritise observations consistent with the video label and note discrepancies as ambiguous cues.
- Focus on concrete forensic signals and visual evidence mentioned by the models.
\end{lstlisting}

\par\smallskip
\textbf{DeepSeek-V4 Pro System Prompt (continued)}
\begin{lstlisting}[style=prompt]
- Do NOT output a real/fake verdict. Only output the observation text.
- Do NOT mention or reference the ground-truth label, the video's real/fake status, or whether the video is real or fake in your output. The observation must read as if the observer does not know the verdict.
- Do NOT wrap your output in any tags. Output plain text only.
\end{lstlisting}
\textbf{DeepSeek-V4 Pro User Prompt Template}
\begin{lstlisting}[style=prompt,belowskip=0pt]
This video is {ground_truth}.
Below are {n} observation reports from forensic models that analyzed the {dimension_name} of this video. Each report focuses on {dimension_description}.

--- Model 1 ({model_name_1}) ---
{observation_1}

...

--- Model {n} ({model_name_n}) ---
{observation_n}
\end{lstlisting}
\end{promptsingle}

\begin{promptsingle}{Prompt shared by DeepSeek-V4 Pro and GPT-5-mini for evaluating the generated forensic explanations, following the explanation evaluation prompt of VidGuard-R1. Each judge returns a concise analysis and a holistic integer score from 1 to 10.}{lst:explanation-quality-evaluator-prompt}
\textbf{Evaluator System Prompt}
\begin{lstlisting}[style=prompt]
You are an independent expert evaluating the quality of an explanation
generated by a face-video forgery detection model. The model predicts whether
a video is real or fake and provides a rationale for its prediction.

Evaluate the rationale with respect to forensic evidence involving texture and
detail, lighting, motion and temporal consistency, and physical plausibility.

Consider the following five criteria:

1. Evidence accuracy:
Does the rationale identify relevant signs of manipulation or authenticity?

2. Reference alignment:
Does it agree with the important visual evidence described in the reference
rationale?

3. Specificity and grounding:
Does it provide concrete, video-related observations rather than vague claims,
unsupported assumptions, or hallucinated details?

\end{lstlisting}

\par\smallskip
\textbf{Evaluator System Prompt (continued)}
\begin{lstlisting}[style=prompt]
4. Clarity and conciseness:
Is the rationale coherent, direct, and free from unnecessary repetition?

5. Verdict consistency:
Does the rationale logically support the model's predicted real/fake label?
A rationale that contradicts the ground-truth label or its own predicted label
should not receive a high score.

Assign one holistic integer score from 1 to 10:
- 1 indicates an incorrect, unsupported, or contradictory rationale.
- 10 indicates an accurate, specific, concise rationale that fully aligns with
  the reference evidence and convincingly supports the correct verdict.

Return exactly two lines:

ANALYSIS: <a concise paragraph evaluating the rationale>
SCORE: <an integer from 1 to 10>

Do not output anything else.
\end{lstlisting}

\textbf{Evaluator User Prompt Template}
\begin{lstlisting}[style=prompt,belowskip=0pt]
Evaluate the following model-generated explanation.

Ground-truth answer:
{ground_truth_answer}

Reference rationale:
{ground_truth_rationale}

Model answer:
{model_answer}

Model rationale:
{model_rationale}
\end{lstlisting}
\end{promptsingle}

\section{Training Dynamics}
\label{sec:training-dynamics}
\begin{figure*}[t]
    \centering
    \begin{minipage}[t]{0.48\textwidth}
        \centering
        \includegraphics[width=\linewidth]{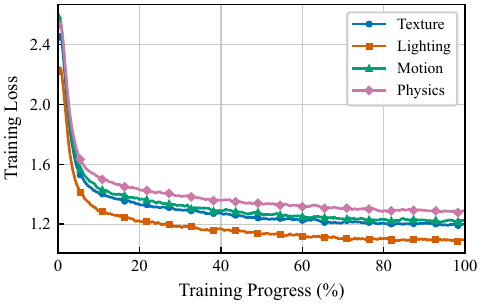}
        {\textbf{(a)} Observation loss\par}
    \end{minipage}
    \hfill
    \begin{minipage}[t]{0.48\textwidth}
        \centering
        \includegraphics[width=\linewidth]{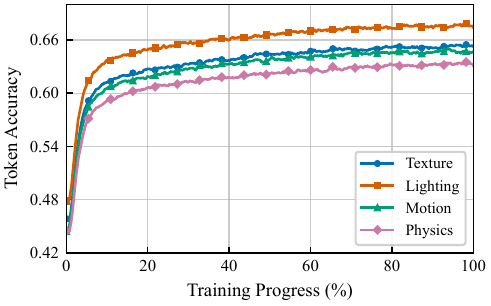}
        {\textbf{(b)} Observation accuracy\par}
    \end{minipage}

    \smallskip

    \begin{minipage}[t]{0.48\textwidth}
        \centering
        \includegraphics[width=\linewidth]{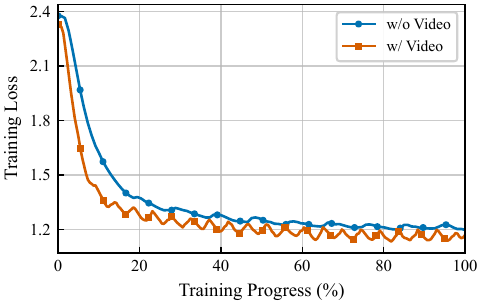}
        {\textbf{(c)} Judge loss\par}
    \end{minipage}
    \hfill
    \begin{minipage}[t]{0.48\textwidth}
        \centering
        \includegraphics[width=\linewidth]{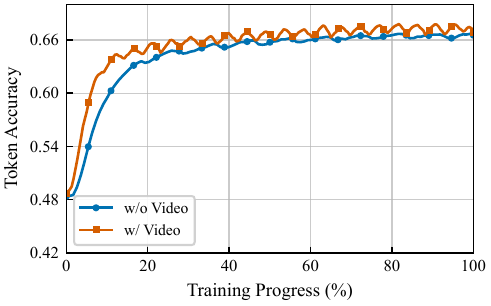}
        {\textbf{(d)} Judge accuracy\par}
    \end{minipage}

    \smallskip

    \begin{minipage}[t]{0.32\textwidth}
        \centering
        \includegraphics[width=\linewidth]{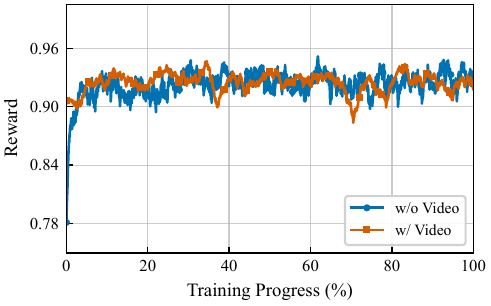}
        {\textbf{(e)} Mean reward\par}
    \end{minipage}
    \hfill
    \begin{minipage}[t]{0.32\textwidth}
        \centering
        \includegraphics[width=\linewidth]{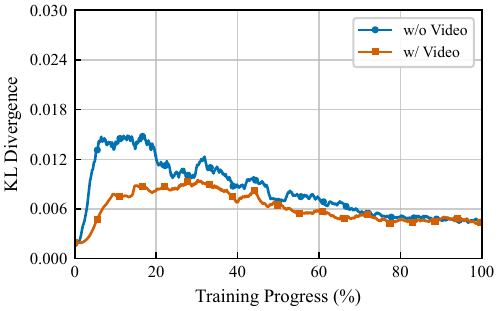}
        {\textbf{(f)} KL divergence\par}
    \end{minipage}
    \hfill
    \begin{minipage}[t]{0.32\textwidth}
        \centering
        \includegraphics[width=\linewidth]{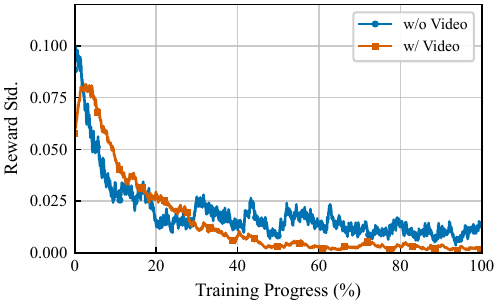}
        {\textbf{(g)} Reward standard deviation\par}
    \end{minipage}
    \caption{Training dynamics of the framework. (a--d) SFT loss and token
    accuracy for the four observation agents and for the judge agent with and
    without video input. (e--g) GRPO mean reward, KL divergence, and reward
    standard deviation for the judge agent in the same two settings.}
    \label{fig:training-dynamics}
\end{figure*}

\Cref{fig:training-dynamics}(a--d) shows the SFT dynamics. All four observation agents converge in the same manner: the training loss drops steeply within roughly the first 10\% of training and then declines gradually, while token accuracy rises to a plateau that is essentially flat over the second half of training. The four agents remain separated by a small and stable margin, with the lighting observation agent attaining the lowest loss and the highest token accuracy and the physics observation agent the highest loss and the lowest token accuracy. The two judge agent configurations converge to nearly the same point, with training losses approaching 1.2 and token accuracies approaching 0.66. They differ mainly in how quickly they get there: with video input the loss falls faster over the first 20\% of training, and its curve remains slightly more oscillatory thereafter, whereas the two configurations are largely indistinguishable once training passes the halfway point.

\Cref{fig:training-dynamics}(e--g) shows the GRPO dynamics of the judge agent. The mean reward rises sharply within early training and fluctuates around a high plateau, with the two configurations closely aligned and neither consistently ahead; because the reward is the binary accuracy of the parsed verdict, this indicates that GRPO refines an already correct decision policy rather than learning it from scratch. The KL divergence against the frozen SFT policy separates the two configurations more clearly: with video input it stays lower through the first 60\% of training, peaking near 0.010 against 0.015 without video input, after which both decay below 0.006. The judge with video input reaches comparable reward while drifting less from its SFT initialization. The reward standard deviation decreases in both configurations as the sampled candidates increasingly agree, and it falls to nearly zero with video input over the final third of training but remains higher without video.

\raggedbottom

\bibliography{arxiv}

@misc{deepfakedetection,
  author       = {Dufour, Nicholas and Gully, Andrew and Karlsson, Per and Vorbyov, Alexey Victor and Leung, Thomas and Childs, Jeremiah and Bregler, Christoph},
  title        = {{DeepFakes Detection Dataset by Google \& Jigsaw}},
  year         = {2019},
  month        = sep,
  note         = {Accessed: 2026-05-11}
}

@misc{dfdc,
  title={The deepfake detection challenge (dfdc) dataset},
  author={Dolhansky, Brian and Bitton, Joanna and Pflaum, Ben and Lu, Jikuo and Howes, Russ and Wang, Menglin and Ferrer, Cristian Canton},
  journal={arXiv preprint arXiv:2006.07397},
  year={2020}
}

@inproceedings{celeb-df,
  title={Celeb-df: A large-scale challenging dataset for deepfake forensics},
  author={Li, Yuezun and Yang, Xin and Sun, Pu and Qi, Honggang and Lyu, Siwei},
  booktitle={Proceedings of the IEEE/CVF conference on computer vision and pattern recognition},
  pages={3207--3216},
  year={2020}
}

@inproceedings{deeperforensics,
  title={Deeperforensics-1.0: A large-scale dataset for real-world face forgery detection},
  author={Jiang, Liming and Li, Ren and Wu, Wayne and Qian, Chen and Loy, Chen Change},
  booktitle={Proceedings of the IEEE/CVF conference on computer vision and pattern recognition},
  pages={2889--2898},
  year={2020}
}

@article{df40,
  title={Df40: Toward next-generation deepfake detection},
  author={Yan, Zhiyuan and Yao, Taiping and Chen, Shen and Zhao, Yandan and Fu, Xinghe and Zhu, Junwei and Luo, Donghao and Wang, Chengjie and Ding, Shouhong and Wu, Yunsheng and others},
  journal={Advances in Neural Information Processing Systems},
  volume={37},
  pages={29387--29434},
  year={2024}
}

@misc{celeb-df++,
  title={Celeb-df++: A large-scale challenging video deepfake benchmark for generalizable forensics},
  author={Li, Yuezun and Zhu, Delong and Cui, Xinjie and Lyu, Siwei},
  journal={arXiv preprint arXiv:2507.18015},
  year={2025}
}

@article{tall++,
  title={Learning Spatiotemporal Inconsistency via Thumbnail Layout for Face Deepfake Detection},
  author={Xu, Yuting and Liang, Jian and Sheng, Lijun and Zhang, Xiao-Yu},
  journal={International Journal of Computer Vision},
  volume={132},
  number={12},
  pages={5663--5680},
  year={2024},
  publisher={Springer US New York}
}

@inproceedings{tfcu,
  title={Face forgery video detection via temporal forgery cue unraveling},
  author={Guo, Zonghui and Liu, Yingjie and Zhang, Jie and Zheng, Haiyong and Shan, Shiguang},
  booktitle={Proceedings of the Computer Vision and Pattern Recognition Conference},
  pages={7396--7405},
  year={2025}
}

@inproceedings{dfd-cfg,
  title={Towards more general video-based deepfake detection through facial component guided adaptation for foundation model},
  author={Han, Yue-Hua and Huang, Tai-Ming and Hua, Kai-Lung and Chen, Jun-Cheng},
  booktitle={Proceedings of the IEEE/CVF conference on computer vision and pattern recognition},
  pages={22995--23005},
  year={2025}
}

@article{dfgaze,
  title={Where deepfakes gaze at? Spatial--temporal gaze inconsistency analysis for video face forgery detection},
  author={Peng, Chunlei and Miao, Zimin and Liu, Decheng and Wang, Nannan and Hu, Ruimin and Gao, Xinbo},
  journal={IEEE Transactions on Information Forensics and Security},
  volume={19},
  pages={4507--4517},
  year={2024},
  publisher={IEEE}
}

@inproceedings{effort,
  title={Orthogonal Subspace Decomposition for Generalizable AI-Generated Image Detection},
  author={Yan, Zhiyuan and Wang, Jiangming and Jin, Peng and Zhang, Ke-Yue and Liu, Chengchun and Chen, Shen and Yao, Taiping and Ding, Shouhong and Wu, Baoyuan and Yuan, Li},
  booktitle={International Conference on Machine Learning},
  pages={70268--70288},
  year={2025},
  organization={PMLR}
}

@inproceedings{skyra,
  title     = {Skyra: AI-Generated Video Detection via Grounded Artifact Reasoning},
  author    = {Li, Yifei and Zheng, Wenzhao and Zhang, Yanran and Sun, Runze and Zheng, Yu and Chen, Lei and Zhou, Jie and Lu, Jiwen},
  booktitle = {Proceedings of the IEEE/CVF Conference on Computer Vision and Pattern Recognition},
  year      = {2026}
}

@misc{gpt4o,
  author       = {{OpenAI}},
  title        = {{GPT-4o System Card}},
  year         = {2024},
  month        = aug,
  howpublished = {\url{https://openai.com/index/gpt-4o-system-card/}},
}

@misc{qwen25vl,
  author       = {Bai, Shuai and Chen, Keqin and Liu, Xuejing and Wang, Jialin and Ge, Wenbin and Song, Sibo and Dang, Kai and Wang, Peng and Wang, Shijie and Tang, Jun and Zhong, Humen and Zhu, Yuanzhi and Yang, Mingkun and Li, Zhaohai and Wan, Jianqiang and Wang, Pengfei and Ding, Wei and Fu, Zheren and Xu, Yiheng and Ye, Jiabo and Zhang, Xi and Xie, Tianbao and Cheng, Zesen and Zhang, Hang and Yang, Zhibo and Xu, Haiyang and Lin, Junyang},
  title        = {{Qwen2.5-VL Technical Report}},
  journal      = {arXiv preprint arXiv:2502.13923},
  year         = {2025},
  eprint       = {2502.13923},
  archivePrefix = {arXiv},
  primaryClass = {cs.CV},
  url          = {https://arxiv.org/abs/2502.13923}
}

@misc{videoveritas,
  author    = {Tan, Hao and Lan, Jun and Shi, Senyuan and Tan, Zichang and Yu, Zijian and Zhu, Huijia and Wang, Weiqiang and Wan, Jun and Lei, Zhen},
  title     = {{VideoVeritas: AI-Generated Video Detection via Perception Pretext Reinforcement Learning}},
  booktitle = {Proceedings of the 43rd International Conference on Machine Learning},
  year      = {2026},
  url       = {https://arxiv.org/abs/2602.08828}
}

@article{genvidbench,
  title={GenVidBench: A 6-Million Benchmark for AI-Generated Video Detection},
  author={Ni, Zhenliang and Yan, Qiangyu and Huang, Mouxiao and Yuan, Tianning and Tang, Yehui and Hu, Hailin and Chen, Xinghao and Wang, Yunhe},
  journal={Proceedings of the AAAI Conference on Artificial Intelligence},
  volume={40},
  number={18},
  pages={15582--15590},
  year={2026},
  doi={10.1609/aaai.v40i18.38587},
  url={https://ojs.aaai.org/index.php/AAAI/article/view/38587}
}

@misc{AIGVDBench,
  title={Your One-Stop Solution for AI-Generated Video Detection},
  author={Ma, Long and Xue, Zihao and Wang, Yan and Yan, Zhiyuan and Xu, Jin and Jiang, Xiaorui and Yu, Haiyang and Liao, Yong and Bi, Zhen},
  booktitle={Proceedings of the IEEE/CVF Conference on Computer Vision and Pattern Recognition},
  year={2026},
  doi={10.48550/arXiv.2601.11035},
  url={https://arxiv.org/abs/2601.11035}
}

@inproceedings{ff++,
  title={Faceforensics++: Learning to detect manipulated facial images},
  author={Rossler, Andreas and Cozzolino, Davide and Verdoliva, Luisa and Riess, Christian and Thies, Justus and Nie{\ss}ner, Matthias},
  booktitle={Proceedings of the IEEE/CVF international conference on computer vision},
  pages={1--11},
  year={2019}
}

@misc{runwaygen3,
  title        = {Introducing {Gen-3 Alpha}: A New Frontier for Video Generation},
  author       = {{Runway Research}},
  howpublished = {\url{https://runwayml.com/research/introducing-gen-3-alpha}},
  year         = {2024}
}

@inproceedings{causvid,
  title     = {From Slow Bidirectional to Fast Autoregressive Video Diffusion Models},
  author    = {Yin, Tianwei and Zhang, Qiang and Zhang, Richard and Freeman, William T. and Durand, Fredo and Shechtman, Eli and Huang, Xun},
  booktitle = {Proceedings of the IEEE/CVF Conference on Computer Vision and Pattern Recognition},
  pages     = {22963--22974},
  year      = {2025}
}

@misc{modelscope,
  title   = {{ModelScope} Text-to-Video Technical Report},
  author  = {Wang, Jiuniu and Yuan, Hangjie and Chen, Dayou and Zhang, Yingya and Wang, Xiang and Zhang, Shiwei},
  journal = {arXiv preprint arXiv:2308.06571},
  year    = {2023},
  doi     = {10.48550/arXiv.2308.06571},
  url     = {https://arxiv.org/abs/2308.06571}
}

@misc{pika,
  title        = {{Pika}},
  author       = {{Pika}},
  howpublished = {\url{https://pika.art/}},
  year         = {2024},
}

@inproceedings{text2videozero,
  title     = {{Text2Video-Zero}: Text-to-Image Diffusion Models are Zero-Shot Video Generators},
  author    = {Khachatryan, Levon and Movsisyan, Andranik and Tadevosyan, Vahram and Henschel, Roberto and Wang, Zhangyang and Navasardyan, Shant and Shi, Humphrey},
  booktitle = {Proceedings of the IEEE/CVF International Conference on Computer Vision},
  pages     = {15954--15964},
  year      = {2023}
}

@inproceedings{videocrafter2,
  title     = {{VideoCrafter2}: Overcoming Data Limitations for High-Quality Video Diffusion Models},
  author    = {Chen, Haoxin and Zhang, Yong and Cun, Xiaodong and Xia, Menghan and Wang, Xintao and Weng, Chao and Shan, Ying},
  booktitle = {Proceedings of the IEEE/CVF Conference on Computer Vision and Pattern Recognition},
  pages     = {7310--7320},
  year      = {2024}
}

@inproceedings{cogvideo,
  title     = {{CogVideo}: Large-scale Pretraining for Text-to-Video Generation via Transformers},
  author    = {Hong, Wenyi and Ding, Ming and Zheng, Wendi and Liu, Xinghan and Tang, Jie},
  booktitle = {International Conference on Learning Representations},
  year      = {2023},
  url       = {https://openreview.net/forum?id=rB6TpjAuSRy}
}

@misc{cogvideox1.5-5b-t,
  title        = {{CogVideoX1.5-5B}},
  author       = {{Z.ai}},
  howpublished = {\url{https://huggingface.co/zai-org/CogVideoX1.5-5B}},
  year         = {2024},
}

@misc{hunyuanvideo,
  title   = {{HunyuanVideo}: A Systematic Framework for Large Video Generative Models},
  author  = {Kong, Weijie and Tian, Qi and Zhang, Zijian and Min, Rox and Dai, Zuozhuo and Zhou, Jin and Xiong, Jiangfeng and Li, Xin and Wu, Bo and Zhang, Jianwei and others},
  journal = {arXiv preprint arXiv:2412.03603},
  year    = {2024},
  doi     = {10.48550/arXiv.2412.03603},
  url     = {https://arxiv.org/abs/2412.03603}
}

@misc{ltx-video-13b-I,
  title   = {{LTX-Video}: Realtime Video Latent Diffusion},
  author  = {HaCohen, Yoav and Chiprut, Nisan and Brazowski, Benny and Shalem, Daniel and Moshe, Dudu and Richardson, Eitan and Levin, Eran and Shiran, Guy and Zabari, Nir and Gordon, Ori and Panet, Poriya and Weissbuch, Sapir and Kulikov, Victor and Bitterman, Yaki and Melumian, Zeev and Bibi, Ofir},
  journal = {arXiv preprint arXiv:2501.00103},
  year    = {2025},
  doi     = {10.48550/arXiv.2501.00103},
  url     = {https://arxiv.org/abs/2501.00103},
}

@misc{skyreels-v2,
  title   = {{SkyReels-V2}: Infinite-length Film Generative Model},
  author  = {Chen, Guibin and Lin, Dixuan and Yang, Jiangping and Lin, Chunze and Zhu, Juncheng and Fan, Mingyuan and Zhang, Hao and others},
  journal = {arXiv preprint arXiv:2504.13074},
  year    = {2025},
  doi     = {10.48550/arXiv.2504.13074},
  url     = {https://arxiv.org/abs/2504.13074}
}

@misc{wan2.1-t2v-1.3b,
  title   = {{Wan}: Open and Advanced Large-Scale Video Generative Models},
  author  = {{Wan Team} and Wang, Ang and Ai, Baole and Wen, Bin and Mao, Chaojie and Xie, Chen-Wei and Chen, Di and others},
  journal = {arXiv preprint arXiv:2503.20314},
  year    = {2025},
  doi     = {10.48550/arXiv.2503.20314},
  url     = {https://arxiv.org/abs/2503.20314},
}

@misc{runwaygen4turbo,
  title        = {Runway {Gen-4}: AI Video Generation with World Consistency},
  author       = {{Runway Research}},
  howpublished = {\url{https://runwayml.com/research/introducing-runway-gen-4}},
  year         = {2025},
}

@misc{hailuo,
  title        = {{Hailuo Video}},
  author       = {{MiniMax}},
  howpublished = {\url{https://www.minimax.io/}},
  year         = {2024}
}

@misc{kling,
  title        = {{Kling AI} Video Generation},
  author       = {{Kuaishou}},
  howpublished = {\url{https://kling.ai/}},
  year         = {2024}
}

@misc{pika2.2,
  title        = {{Pika} Model 2.2},
  author       = {{Pika}},
  howpublished = {\url{https://pika.art/faq}},
  year         = {2025}
}

@misc{pixverse-v4.5,
  title        = {{PixVerse v4.5}},
  author       = {{PixVerse}},
  howpublished = {\url{https://pixverse.ai/en}},
  year         = {2025}
}

@misc{sora2,
  title        = {{Sora 2} System Card},
  author       = {{OpenAI}},
  howpublished = {\url{https://openai.com/index/sora-2-system-card/}},
  year         = {2025}
}

@misc{seedance2.0,
  title   = {{Seedance 2.0}: Advancing Video Generation for World Complexity},
  author  = {{Team Seedance} and Chen, De and Chen, Liyang and Chen, Xin and Chen, Ying and Chen, Zhuo and others},
  journal = {arXiv preprint arXiv:2604.14148},
  year    = {2026},
  doi     = {10.48550/arXiv.2604.14148},
  url     = {https://arxiv.org/abs/2604.14148}
}

@inproceedings{altfreezing,
  title={Altfreezing for more general video face forgery detection},
  author={Wang, Zhendong and Bao, Jianmin and Zhou, Wengang and Wang, Weilun and Li, Houqiang},
  booktitle={Proceedings of the IEEE/CVF conference on computer vision and pattern recognition},
  pages={4129--4138},
  year={2023}
}

@misc{qwen3.6-35b-a3b,
    title = {{Qwen3.6-35B-A3B}: Agentic Coding Power, Now Open to All},
    author = {{Qwen Team}},
    year = {2026},
    month = {April},
    url = {https://qwen.ai/blog?id=qwen3.6-35b-a3b}
}

@misc{mimov25,
  title        = {{MiMo-V2.5}},
  author       = {{Xiaomi MiMo Team}},
  year         = {2026},
  howpublished = {\url{https://huggingface.co/collections/XiaomiMiMo/mimo-v25}},
}

@inproceedings{Unishield,
  title={Unishield: An adaptive multi-agent framework for unified forgery image detection and localization},
  author={Huang, Qing and Xu, Zhipei and Zhang, Xuanyu and Yu, Xiangyu and Zhang, Jian},
  booktitle={Proceedings of the IEEE/CVF Conference on Computer Vision and Pattern Recognition},
  pages={8121--8132},
  year={2026}
}

@inproceedings{longvideoagent,
    title = "{L}ong{V}ideo{A}gent: Multi-Agent Reasoning with Long Videos",
    author = "Liu, Runtao  and
      Liu, Ziyi  and
      Tang, Jiaqi  and
      Ma, Yue  and
      Pi, Renjie  and
      Zhang, Jipeng  and
      Chen, Qifeng",
    editor = "Liakata, Maria  and
      Moreira, Viviane P.  and
      Zhang, Jiajun  and
      Jurgens, David",
    booktitle = "Proceedings of the 64th Annual Meeting of the {A}ssociation for {C}omputational {L}inguistics (Volume 1: Long Papers)",
    month = jul,
    year = "2026",
    address = "San Diego, California, United States",
    publisher = "Association for Computational Linguistics",
    url = "https://aclanthology.org/2026.acl-long.1876/",
    doi = "10.18653/v1/2026.acl-long.1876",
    pages = "40404--40416",
    ISBN = "979-8-89176-390-6"
}

@misc{gao2025singleagentmultiagentsystemsboth,
      title={Single-agent or Multi-agent Systems? Why Not Both?}, 
      author={Mingyan Gao and Yanzi Li and Banruo Liu and Yifan Yu and Phillip Wang and Ching-Yu Lin and Fan Lai},
      year={2025},
      eprint={2505.18286},
      archivePrefix={arXiv},
      primaryClass={cs.MA},
      url={https://arxiv.org/abs/2505.18286}
}

@inproceedings{MARPO,
  title={MARPO: A Reflective Policy Optimization for Multi-Agent Reinforcement Learning},
  author={Wu, Cuiling and Gan, Yaozhong and Xing, Junliang and Fu, Ying},
  booktitle={Proceedings of the AAAI Conference on Artificial Intelligence},
  volume={40},
  pages={29740--29748},
  year={2026}
}

@misc{he2025enhancingllmreasoningmultipath,
      title={Enhancing LLM Reasoning with Multi-Path Collaborative Reactive and Reflection agents}, 
      author={Chengbo He and Bochao Zou and Xin Li and Jiansheng Chen and Junliang Xing and Huimin Ma},
      year={2025},
      eprint={2501.00430},
      archivePrefix={arXiv},
      primaryClass={cs.CL},
      url={https://arxiv.org/abs/2501.00430}
}

@ARTICLE{face-mogle,
  author={Zou, Xuechao and Zhang, Shun and Fu, Xing and Li, Yue and Li, Kai and Cao, Yushe and Lang, Congyan and Tao, Pin and Xing, Junliang},
  journal={IEEE Transactions on Pattern Analysis and Machine Intelligence}, 
  title={Mixture of Global and Local Experts with Diffusion Transformer for Controllable Face Generation}, 
  year={2026},
  volume={},
  number={},
  pages={1-17},
  doi={10.1109/TPAMI.2026.3708691}
}

@inproceedings{cao2026multivariate,
  title={Multivariate diffusion transformer with decoupled attention for high-fidelity mask-text collaborative facial generation},
  author={Cao, Yushe and Shi, Dianxi and Fu, Xing and Zou, Xuechao and Peng, Haikuo and Li, Xueqi and Yu, Chun and Xing, Junliang},
  booktitle={Proceedings of the AAAI Conference on Artificial Intelligence},
  volume={40},
  pages={2670--2679},
  year={2026}
}

@inproceedings{adamw,
  title={Decoupled Weight Decay Regularization},
  author={Loshchilov, Ilya and Hutter, Frank},
  booktitle={ICLR},
  year={2018}
}

@inproceedings{internvl,
  title={Intern VL: Scaling up Vision Foundation Models and Aligning for Generic Visual-Linguistic Tasks},
  author={Chen, Zhe and Wu, Jiannan and Wang, Wenhai and Su, Weijie and Chen, Guo and Xing, Sen and Zhong, Muyan and Zhang, Qinglong and Zhu, Xizhou and Lu, Lewei and others},
  booktitle={2024 IEEE/CVF Conference on Computer Vision and Pattern Recognition (CVPR)},
  pages={24185--24198},
  year={2024},
  organization={IEEE Computer Society}
}

@inproceedings{
du2024improving,
title={Improving Factuality and Reasoning in Language Models through Multiagent Debate},
author={Yilun Du and Shuang Li and Antonio Torralba and Joshua B. Tenenbaum and Igor Mordatch},
booktitle={Forty-first International Conference on Machine Learning},
year={2024},
url={https://openreview.net/forum?id=zj7YuTE4t8}
}

@inproceedings{wu2024autogen,
title={AutoGen: Enabling Next-Gen {LLM} Applications via Multi-Agent Conversations},
author={Qingyun Wu and Gagan Bansal and Jieyu Zhang and Yiran Wu and Beibin Li and Erkang Zhu and Li Jiang and Xiaoyun Zhang and Shaokun Zhang and Jiale Liu and Ahmed Hassan Awadallah and Ryen W White and Doug Burger and Chi Wang},
booktitle={First Conference on Language Modeling},
year={2024},
url={https://openreview.net/forum?id=BAakY1hNKS}
}

@INPROCEEDINGS{Face-X-Ray,
  author={Li, Lingzhi and Bao, Jianmin and Zhang, Ting and Yang, Hao and Chen, Dong and Wen, Fang and Guo, Baining},
  booktitle={2020 IEEE/CVF Conference on Computer Vision and Pattern Recognition (CVPR)}, 
  title={Face X-Ray for More General Face Forgery Detection}, 
  year={2020},
  volume={},
  number={},
  pages={5000-5009},
  doi={10.1109/CVPR42600.2020.00505},
  ISSN={2575-7075},
  month={June},}

@INPROCEEDINGS{Lips-Donot-Lie,
  author={Haliassos, Alexandros and Vougioukas, Konstantinos and Petridis, Stavros and Pantic, Maja},
  booktitle={2021 IEEE/CVF Conference on Computer Vision and Pattern Recognition (CVPR)}, 
  title={Lips Don't Lie: A Generalisable and Robust Approach to Face Forgery Detection}, 
  year={2021},
  volume={},
  number={},
  pages={5037-5047},
  doi={10.1109/CVPR46437.2021.00500},
  ISSN={2575-7075},
  month={June},}

@INPROCEEDINGS{9710282,
  author={Zheng, Yinglin and Bao, Jianmin and Chen, Dong and Zeng, Ming and Wen, Fang},
  booktitle={2021 IEEE/CVF International Conference on Computer Vision (ICCV)}, 
  title={Exploring Temporal Coherence for More General Video Face Forgery Detection}, 
  year={2021},
  volume={},
  number={},
  pages={15024-15034},
  doi={10.1109/ICCV48922.2021.01477},
  ISSN={2380-7504},
  month={Oct},}

@inproceedings{drmas,
title={Dr. {MAS}: Stable Reinforcement Learning for Multi-Agent {LLM} Systems},
author={Lang Feng and Longtao Zheng and Shuo He and Fuxiang Zhang and Bo An},
booktitle={Workshop on Multi-Agent Learning and Its Opportunities in the Era of Generative AI},
year={2026},
url={https://openreview.net/forum?id=GOX1vk1IL1}
}

@inproceedings{
PettingLLMs,
title={Stronger-{MAS}: Multi-Agent Reinforcement Learning for Collaborative {LLM}s},
author={Yujie Zhao and Lanxiang Hu and Yang Wang and Minmin Hou and Hao Zhang and Ke Ding and Jishen Zhao},
booktitle={The Fourteenth International Conference on Learning Representations},
year={2026},
url={https://openreview.net/forum?id=IdF6JqXWzx}
}

@inproceedings{VidGuard-R1,
title={VidGuard-R1: {AI}-Generated Video Detection and Explanation via Reasoning {MLLM}s and {RL}},
author={Kyoungjun Park and Yifan Yang and Juheon Yi and Muhammad Muaz and Shicheng Zheng and Yifei Shen and Dongqi Han and Caihua Shan and Lili Qiu},
booktitle={The Fourteenth International Conference on Learning Representations},
year={2026},
url={https://openreview.net/forum?id=gXjOsBcXIR}
}

@inproceedings{thies2016face2face,
  title={Face2face: Real-time face capture and reenactment of rgb videos},
  author={Thies, Justus and Zollhofer, Michael and Stamminger, Marc and Theobalt, Christian and Nie{\ss}ner, Matthias},
  booktitle={Proceedings of the IEEE conference on computer vision and pattern recognition},
  pages={2387--2395},
  year={2016}
}

@article{NeuralTextures,
  title={Deferred neural rendering: Image synthesis using neural textures},
  author={Thies, Justus and Zollh{\"o}fer, Michael and Nie{\ss}ner, Matthias},
  journal={Acm Transactions on Graphics (TOG)},
  volume={38},
  number={4},
  pages={1--12},
  year={2019},
  publisher={ACM New York, NY, USA}
}

@inproceedings{FaceShifter,
  title={Advancing high fidelity identity swapping for forgery detection},
  author={Li, Lingzhi and Bao, Jianmin and Yang, Hao and Chen, Dong and Wen, Fang},
  booktitle={Proceedings of the IEEE/CVF conference on computer vision and pattern recognition},
  pages={5074--5083},
  year={2020}
}

@misc{EDVD-LLaMA,
  title={EDVD-LLaMA: Explainable Deepfake Video Detection via Multimodal Large Language Model Reasoning},
  author={Sun, Haoran and Cai, Chen and Zhuang, Huiping and Lee, Kong Aik and Chau, Lap-Pui and Wang, Yi},
  journal={arXiv preprint arXiv:2510.16442},
  year={2025}
}

@misc{qiao2026offline,
title={Offline Multi-Agent Reinforcement Learning via Sequential Score Decomposition},
author={Dan Qiao and Wenhao Li and Shanchao Yang and Hongyuan Zha and Baoxiang Wang},
year={2026},
url={https://openreview.net/forum?id=LRu30T4Vev}
}

@misc{deepseek-v4,
  title={Deepseek-v4: Towards highly efficient million-token context intelligence},
  author={{DeepSeek-AI} and Xu, Anyi and Lin, Bangcai and Xue, Bing and Wang, Bingxuan and Xu, Bingzheng and Wu, Bochao and Zhang, Bowei and Lin, Chaofan and Dong, Chen and Ling, Chenchen and others},
  journal={arXiv preprint arXiv:2606.19348},
  year={2026}
}

@misc{gemini35flash,
  author       = {{Google DeepMind}},
  title        = {Gemini 3.5 Flash: Model Card},
  year         = {2026},
  month        = may,
  howpublished = {\url{https://deepmind.google/models/model-cards/gemini-3-5-flash/}},
}

@misc{grpo,
  title={Deepseekmath: Pushing the limits of mathematical reasoning in open language models},
  author={Shao, Zhihong and Wang, Peiyi and Zhu, Qihao and Xu, Runxin and Song, Junxiao and Bi, Xiao and Zhang, Haowei and Zhang, Mingchuan and Li, YK and Wu, Yang and others},
  journal={arXiv preprint arXiv:2402.03300},
  year={2024}
}

@inproceedings{ms-swift,
  title={Swift: a scalable lightweight infrastructure for fine-tuning},
  author={Zhao, Yuze and Huang, Jintao and Hu, Jinghan and Wang, Xingjun and Mao, Yunlin and Zhang, Daoze and Jiang, Zeyinzi and Wu, Zhikai and Ai, Baole and Wang, Ang and others},
  booktitle={Proceedings of the AAAI Conference on Artificial Intelligence},
  volume={39},
  pages={29733--29735},
  year={2025}
}

@inproceedings{lora,
title={Lo{RA}: Low-Rank Adaptation of Large Language Models},
author={Edward J Hu and Yelong Shen and Phillip Wallis and Zeyuan Allen-Zhu and Yuanzhi Li and Shean Wang and Lu Wang and Weizhu Chen},
booktitle={ICLR},
year={2022},
pages={1--13}
}

@techreport{openai2025gpt5,
  title       = {{GPT-5 System Card}},
  author      = {{OpenAI}},
  institution = {OpenAI},
  year        = {2025},
  month       = aug,
  url         = {https://cdn.openai.com/gpt-5-system-card.pdf},
}

@InProceedings{Cheng_2026_CVPR,
    author    = {Cheng, Jikang and Yan, Renye and Yan, Zhiyuan and Gan, Yaozhong and Zhang, Xueyi and Wang, Zhongyuan and Peng, Wei and Liang, Ling},
    title     = {A Sanity Check for Multi-In-Domain Face Forgery Detection in the Real World},
    booktitle = {Proceedings of the IEEE/CVF Conference on Computer Vision and Pattern Recognition (CVPR)},
    month     = {June},
    year      = {2026},
    pages     = {21306-21315},
}

@misc{shen2025generative,
  title={When Generative Replay Meets Evolving Deepfakes: Domain-Aware Relative Weighting for Incremental Face Forgery Detection},
  author={Shen, Hao and Cheng, Jikang and Yan, Renye and Wang, Zhongyuan and Peng, Wei and Huang, Baojin},
  journal={arXiv preprint arXiv:2511.18436},
  year={2025}
}

@misc{cheng2026infodense,
  title={InfoDense: Density-Aware Regional Decisive Replay for Memory-Efficient Incremental Face Forgery Detection},
  author={Cheng, Jikang and Shen, Hao and Zhang, Xueyi and Wang, Guangcheng and Wang, Zhongyuan and Yan, Renye and Huang, Baojin},
  journal={arXiv preprint arXiv:2607.16873},
  year={2026}
}

@inproceedings{li2024safeear,
  title={Safeear: Content privacy-preserving audio deepfake detection},
  author={Li, Xinfeng and Li, Kai and Zheng, Yifan and Yan, Chen and Ji, Xiaoyu and Xu, Wenyuan},
  booktitle={Proceedings of the 2024 on ACM SIGSAC Conference on Computer and Communications Security},
  pages={3585--3599},
  year={2024}
}

@inproceedings{li2026audiotrust,
  title={Audiotrust: Benchmarking the multifaceted trustworthiness of audio large language models},
  author={Li, Kai and Shen, Can and Liu, Yile and Han, Jirui and Zou, Xuechao and WANG, Lionel and Zhang, Shun and Du, Xingjian and Luo, Hanjun and Jin, Yingbin and others},
  booktitle={International Conference on Learning Representations},
  volume={2026},
  pages={24938--25016},
  year={2026}
}

@article{ma2026mmar,
  title={Mmar: A challenging benchmark for deep reasoning in speech, audio, music, and their mix},
  author={Ma, Ziyang and Ma, Yinghao and Zhu, Yanqiao and Yang, Chen and Chao, Yi-Wen and Xu, Ruiyang and Chen, Wenxi and Chen, Yuanzhe and Chen, Zhuo and Cong, Jian and others},
  journal={Advances in Neural Information Processing Systems},
  volume={38},
  year={2026}
}

@InProceedings{Huang_2026_CVPR,
    author    = {Huang, Yihuan and Xue, Jun and Jiajun, Liu and Li, Daixian and Zhang, Tong and Yi, Zhuolin and Ren, Yanzhen and Li, Kai},
    title     = {When AVSR Meets Video Conferencing: Dataset, Degradation, and the Hidden Mechanism Behind Performance Collapse},
    booktitle = {Proceedings of the IEEE/CVF Conference on Computer Vision and Pattern Recognition (CVPR)},
    month     = {June},
    year      = {2026},
    pages     = {4448-4457}
}

@misc{hu2026beavertrainingfreehierarchicalprompt,
      title={BEAVER: A Training-Free Hierarchical Prompt Compression Method via Structure-Aware Page Selection}, 
      author={Zhengpei Hu and Kai Li and Dapeng Fu and Chang Zeng and Yue Li and Yuanhao Tang and Jianqiang Huang},
      year={2026},
      eprint={2603.19635},
      archivePrefix={arXiv},
      primaryClass={cs.CL},
      url={https://arxiv.org/abs/2603.19635}, 
}

\end{document}